\documentclass[lettersize,journal]{IEEEtran}
\usepackage{amsmath,amsfonts}
\usepackage{algorithm, algorithmicx, algpseudocode}
\usepackage{array}

\usepackage{mathrsfs}
\usepackage{stfloats}

\usepackage{booktabs}
\usepackage{multirow}
\usepackage{longtable}
\usepackage {rotating} 
\usepackage{subfigure}
\usepackage[caption=false,font=normalsize,labelfont=sf,textfont=sf]{subfig}
\usepackage{textcomp}
\usepackage{stfloats}
\usepackage{url}
\usepackage{verbatim}
\usepackage{graphicx}
\usepackage{balance}
\usepackage{cite}
\usepackage{cuted}
\def\BibTeX{{\rm B\kern-.05em{\sc i\kern-.025em b}\kern-.08em
		T\kern-.1667em\lower.7ex\hbox{E}\kern-.125emX}}
	
\begin{document}
	
	\title{A Dual-Channel Particle Swarm Optimization Algorithm Based on Adaptive Balance Search}
	\author{
\textbf{		This work has been submitted to the IEEE for possible publication. Copyright may be transferred without notice, after which this version may no longer be accessible.}\\
		Zhenxing Zhang, Tianxian Zhang Xiangliang Xu, Lingjiang Kong, \\

}

	\markboth{Journal of \LaTeX\ Class Files,~Vol.~18, No.~9, April~2024}%
	{Shell \MakeLowercase{\textit{et al.}}: A Sample Article Using IEEEtran.cls for IEEE Journals}
	
	
	\maketitle
	
	\begin{abstract}
		
		The balance between exploration (\textit{Er}) and exploitation (\textit{Ei})  determines the generalization performance of the particle swarm optimization (PSO) algorithm on different problems. Although the insufficient balance  caused by  global best being located near a local minimum has been widely researched, few scholars have systematically paid attention to two behaviors about personal best position (\textbf{\textit{P}}) and global best position (\textbf{\textit{G}}) existing in PSO:  \textbf{\textit{P}}'s uncontrollable-exploitation  and involuntary-exploration guidance behavior, and \textbf{\textit{G}}'s full-time and global guidance behavior,  each of which negatively affects the balance of \textit{Er} and \textit{Ei}. 
		
		With regards to this, we firstly discuss the two behaviors, unveiling the  mechanisms by which they affect the balance, and further pinpoint three key points  for better balancing   \textit{Er} and \textit{Ei}:   eliminating the coupling
		between \textit{\textbf{P}} and \textit{\textbf{G}}, empowering \textit{\textbf{P}} with controllable-exploitation and voluntary-exploration guidance behavior,  controlling  \textbf{\textit{G}}'s full-time and global guidance behavior. Then, we  present a dual-channel PSO algorithm based on adaptive balance search (DCPSO-ABS). This algorithm entails a dual-channel framework to  mitigate the interaction of \textit{\textbf{P}} and \textit{\textbf{G}}, aiding in regulating the behaviors of \textit{\textbf{P}} and \textit{\textbf{G}}, and meanwhile an adaptive balance search strategy for empowering \textit{\textbf{P}} with voluntary-exploration and controllable-exploitation guidance behavior  as well as adaptively controlling \textit{\textbf{G}}’s full-time and global guidance behavior. Finally, three kinds of experiments on 57 benchmark functions are designed to demonstrate that our proposed algorithm has stronger  generalization performance than selected state-of-the-art algorithms.
		
	\end{abstract}
	
	
	\section{Introduction}
	\IEEEPARstart{M}{any} real-world problems are characterized as multimodal, non-convex, non-differentiable,  discontinuous, even black-box problems. To solve such problems with limited computational resources, researchers have developed various population-based optimization algorithms such as particle swarm optimization (PSO) algorithm \cite{kennedy1995particle}, genetic algorithm (GA) \cite{katoch2021review}, differential evolution (DE) \cite{zhou2019self}, etc. For these population-based optimization algorithms, exploration (\textit{Er}) and exploitation (\textit{Ei})  are two cornerstones to help find  optimal solution of a problem \cite{lynn2015heterogeneous}. \textit{Er}, global search, aims at finding promising regions of search space,  and \textit{Ei}, local search, aims at fine-tuned searching to find the optimal solution within one identified promising region.  However,  there is contradiction between \textit{Er} and \textit{Ei}\cite{lynn2015heterogeneous}: overemphasizing on \textit{Er} wastes computational resources to search for inferior regions of search space, slowing down convergence, which naturally has negative impact on \textit{Ei}, and in contrast, overemphasizing on \textit{Ei} can lead to a loss of diversity in the early iteration stages, possibly falling into one local optimal region, which is detrimental to \textit{Er}. Therefore, achieving balance between \textit{Er} and \textit{Ei} is essential for population-based optimization algorithms to effectively solve various problems. 

	Among population-based algorithms, PSO  has been widely applied to various real-world fields \cite{ houssein2021major}, \cite{isho2020persistence} due to its simple optimization mechanism, although it also suffers from the contradiction between between \textit{Er} and \textit{Ei}.

	whose core principle is that each particle is guided by its own previous velocitie, its personal best position ($\boldsymbol{P}$), and current global best position (\textbf{\textit{G}}) to search for global optimum within entire search space. For this reason, it is more particularly crucial for PSO to achieve  harmonious balance between \textit{Er} and \textit{Ei}, so as to achieve higher generalization performance  for solving different problems.

	With regards to this, numerous improved-PSO algorithms have been introduced, which can be widely divided into three categories. The first category is parameter tuning-based PSOs. For example, the  linear decreasing inertia weight \cite{shi1999empirical}, time varying acceleration coefficients \cite{ratnaweera2004self}, and sigmoid-function-based acceleration coefficients \cite{liu2019novel}. The second category is topological structures-based PSOs, such as LIPS \cite{qu2012distance} and DNSPSO \cite{zeng2020dynamic}. Besides, researchers have implicitly constructed topologies using exemplars generated by diverse learning strategies, such as example-based learning strategy \cite{huang2012example}, adaptive learning strategy \cite{wang2018hybrid}, and scatter learning strategy \cite{ren2013scatter}.  The third category is hybridization-based PSOs, such as, hybridization of GWO and PSO \cite{shaheen2021novel}, and hybridization of GA and PSO \cite{7271066}. Despite these advancements, few existing works have explicitly pointed out two critical behaviors existing in \textit{\textbf{P}} and \textit{\textbf{G}} that directly affect on balancing \textit{Er} and \textit{Ei}: \textbf{1) \textbf{\textit{P}}'s uncontrollable-exploitation  and involuntary-exploration guidance behavior}, and \textbf{2)  \textbf{\textit{G}}'s full-time and global guidance behavior}, each of which negatively affects the generalization performance of PSO.
	
\begin{itemize}
	\item{\textit{\textbf{\textit{P}}'s uncontrollable-exploitation  and involuntary-exploration guidance behavior}. In standard PSO,  when \textbf{\textit{G}} is not updated by other particles,  \textit{\textbf{P}} will continuously guide one particle to exploit around the previous search region regardless of the quality of the region,  showing uncontrollable-exploitation guidance behavior. It can lead to exploitation inefficiencies where one particle costs less time on one superior region or more time on inferior one,  impacting the balance of \textit{Er} and \textit{Ei}. Conversely, when \textbf{\textit{G}} is updated by other particles, \textit{\textbf{P}}, under the influence of \textit{\textbf{G}}, passively guides the particle to slowly jump out of the previous search region for other unknown regions. This  phenomenon displays \textit{\textbf{P}}'s involuntary-exploration guidance behavior, implying that being lack of the direction elements from other particles, \textit{\textbf{P}} do not possess voluntary-exploration ability,  leading to a negative effect on balance of \textit{Er} and \textit{Ei}.}

	\item{\textit{\textit{\textbf{G}}'s full-time and global guidance behavior.} From the whole iteration process, each particle is always guided by \textbf{\textit{G}} to move toward the region where \textit{\textbf{G}} is located, which indicates the full-time guidance behavior of \textbf{\textit{G}}. It may cause exploration inefficiencies where one particle exploring an inferior region guided by poor \textbf{\textit{G}} in pre-iteration period (especially in multimodal problems), impacting the balance of \textit{Er} and \textit{Ei}. Furthermore, on each  iteration slice, \textit{\textbf{G}} guides all particles to cluster towards the region where \textit{\textbf{G}} is located, developing the global guidance behavior of \textbf{\textit{G}}.  It can lead to a rapid loss of diversity of population when \textbf{\textit{G}} is located at the local optimal position, consequently weakening exploration ability and negatively affecting the balance of \textit{Er} and \textit{Ei}.}
\end{itemize}

Furthermore, it is worth noting that \textit{\textbf{P}} and \textit{\textbf{G}} are coupled together to  control one particle's state under the influence of single iterative channel and single population in PSO, making it more difficult to achieve a harmonious balance between \textit{Er} and \textit{Ei}. Hence, in order to more  effectively regulate the behaviors of  \textit{\textbf{P}} and \textit{\textbf{G}},  preemptively decoupling \textit{\textbf{P}} and \textit{\textbf{G}} is not an ineffective mean. 

Through the above analysis, the key points of better
balancing \textit{Er} and \textit{Ei} of PSO lie in:\textbf{1) eliminating the coupling between \textbf{\textit{P}} and \textbf{\textit{G}} as much as possible}, \textbf{2) empowering \textbf{\textit{P}} with controllable-exploitation and voluntary-exploration guidance behavior}, 
\textbf{3) adaptively controlling \textit{\textbf{G}}'s full-time and global guidance behavior while maintaining the accelerated convergence ability of \textbf{\textit{G}}}.

In response, introducing two distinct iterative channels—one serving as the standard PSO iterative channel and the other devoid of \textbf{\textit{G}}'s influence, and two rarely interacting classes of particles within the two distinct channels respectively can help decouple between \textbf{\textit{P}} and \textbf{\textit{G}}, also positively aid to manage the behaviors of them. Then, evaluating the quality of the current search region based on the feedback information derived from evolutionary environment  and integrating  other particles' information into the current \textbf{\textit{P}} can aid in empowering \textit{\textbf{P}}'s controllable-exploitation and voluntary-exploration guidance behavior. Moreover, based on the two introduced channels, by constructing a reasonable relation between the number of two channels employed in the iterative process  and the number of iterations, it becomes possible to adaptively control \textbf{\textit{G}}'s full-time and global guidance behavior.

Motivated by the above discussions, our aim is to  put forward a dual-channel PSO  based on adaptive balance search (DCPSO-ABS)  algorithm with the aim of better balancing the  \textit{Er} and \textit{Ei} of the PSO algorithm. The key contributions of this study can be summarized as follows.

\begin{enumerate}{}{}
	\item{ We discover two behaviors existing in PSO:  \textit{\textbf{P}}'s uncontrollable-exploitation  and involuntary-exploration guidance behavior, \textit{\textbf{G}}'s full-time and global guidance behavior, and delve into the underlying reasons causing the two behaviors. Moreover, we also elucidate the intricate mechanism by which the two bahaviors affect the balance between \textit{Er} and \textit{Ei}.}
	
	\item{We propose a dual-channel framework (DC framework), in which constructing two iterative channels including \textit{non-G-channel} with no \textit{\textbf{G}} and \textit{G-channel} with \textit{\textbf{G}}, and further forming two classes of particles—one operating within \textit{non-G-channel} and one within \textit{G-channel}— inspired by cell division, for  disengagement from the  status quo of coupling between \textit{\textbf{P}} and \textit{\textbf{G}} caused by single iterative channel and single population of PSO, thereby contributing to conveniently regulate the behaviors of them.}
	
	\item{We propose an adaptive balance search (ABS) strategy.  In ABS strategy,  we integrate all \textbf{\textit{P}}s direction information in each dimension into current \textbf{\textit{P}} as much as possible for empowering \textbf{\textit{P}} with voluntary-exploration guidance ability. Moreover, we evaluate the quality of the current search region according to the particles' updated states, and establish the mathematic relationship between the number of exploitation in the current search region and  the quality of it to empower \textbf{\textit{P}} with controllable-exploitation guidance ability. Additionally, we introduce the weak-serial relationship between the \textit{non-G} and \textit{G} channels, and establish the mathematic relationship between the number of function evaluations (the number of iterations) and the proportion of the two iterative channels employed,  allowing us to adaptively control \textit{\textbf{G}}'s full-time and global guidance behavior.}
\end{enumerate}

The remainder of this paper is organized as follows. Section \ref{PSO} gives the concept of standard PSO, and reviews variants of PSO. Section \ref{Double-edged sword features of PSO} describes and analyses the bahaviors of \textbf{\textit{P}} and \textit{\textbf{G}} in standard PSO in detail. Section \ref{DCPSO-ABS} introduces DCPSO-ABS algorithm and conducts a comprehensive analysis of its associated capabilities. Section\ref{experiment} encompasses the experiments along with discussions. Conclusion is drawn in Section \ref{conclusion}.
	
	\section{Review Work}\label{PSO}
	Since PSO was first proposed \cite{kennedy1995particle}, many works have been done to develop it. In this section,  standard PSO proposed by \textit{Shi} and  \textit{Eberhart} \cite{shi1998modified},   and variants of PSO are reviewed successively, regarded as the research background of this paper.
	
	\subsection{ Standard Particle Swarm Optimization}
 In standard PSO,  each particle  treated as  a potential solution flies within search space for optimal solution. 	For a $D$-dimensional optimization problem, the flight of $n$th particle is controlled by velocity  $\boldsymbol{V_n }= (v^1_n,v^2_n,...,v^D_n )$, position $ \boldsymbol{X_n} = (x^1_n,x^2_n,...,x^D_n)$, personal best position $\boldsymbol{P_n} = (p^1_n,p^2_n,...,p^D_n )$ and global best position $ \boldsymbol{G}=(g^1,g^2,...,g^D)$, which can be specifically described by the following equation
	
	\begin{equation} \label{condition pso equat 1}
		\begin{cases}
			{v}^d_n = w\cdot {v}^d_n	+{c_1}\cdot {r_1} \cdot ({p}^d_n-{x}^d_n)	+{c_2}\cdot {r_2} \cdot (g^d-{x}^d_n)\\ 
			{x}^d_n={x}^d_n+{v}^d_n
		\end{cases}
	\end{equation}
	where $w$ represents the inertia weight,   $ {c_1}$ and $ {c_2}$ represent the cognitive and social acceleration coefficients respecitively, $r_1$ and $r_2$ are two random numbers selected evenly within [0,1]. 
	\subsection{Variants of PSO}\label{variants of PSO}
	One of the primary obstacles encountered in PSO pertains to achieving a harmonious balance between the \textit{Er} and \textit{Ei}. In response, researchers have put forth a great variety of   PSO variants, whose core ideal can be broadly classified into  three distinct categories: parameter tuning, population topology construction, and hybridization.  A review of these categories is provided below.
	
	\begin{list}{}
		\item{ \emph{1) Parameter tuning }}
	\end{list}
	
	 \begin{figure*}	
		\subfigure[]{		
			\includegraphics[width=1.4in]{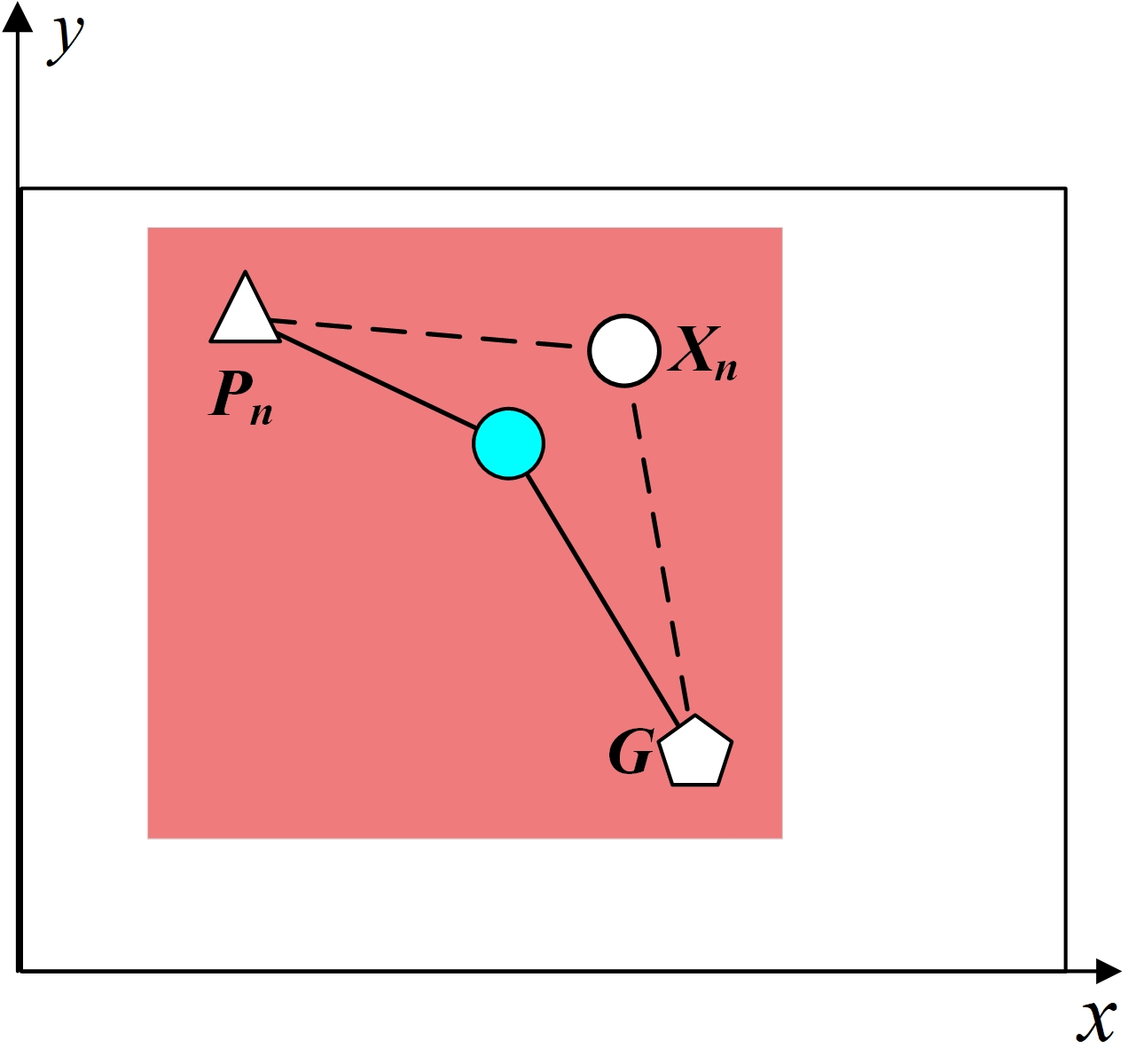}\\
		}%
		\hspace{-2.5mm}
		\subfigure[]{
			\includegraphics[width=1.4in]{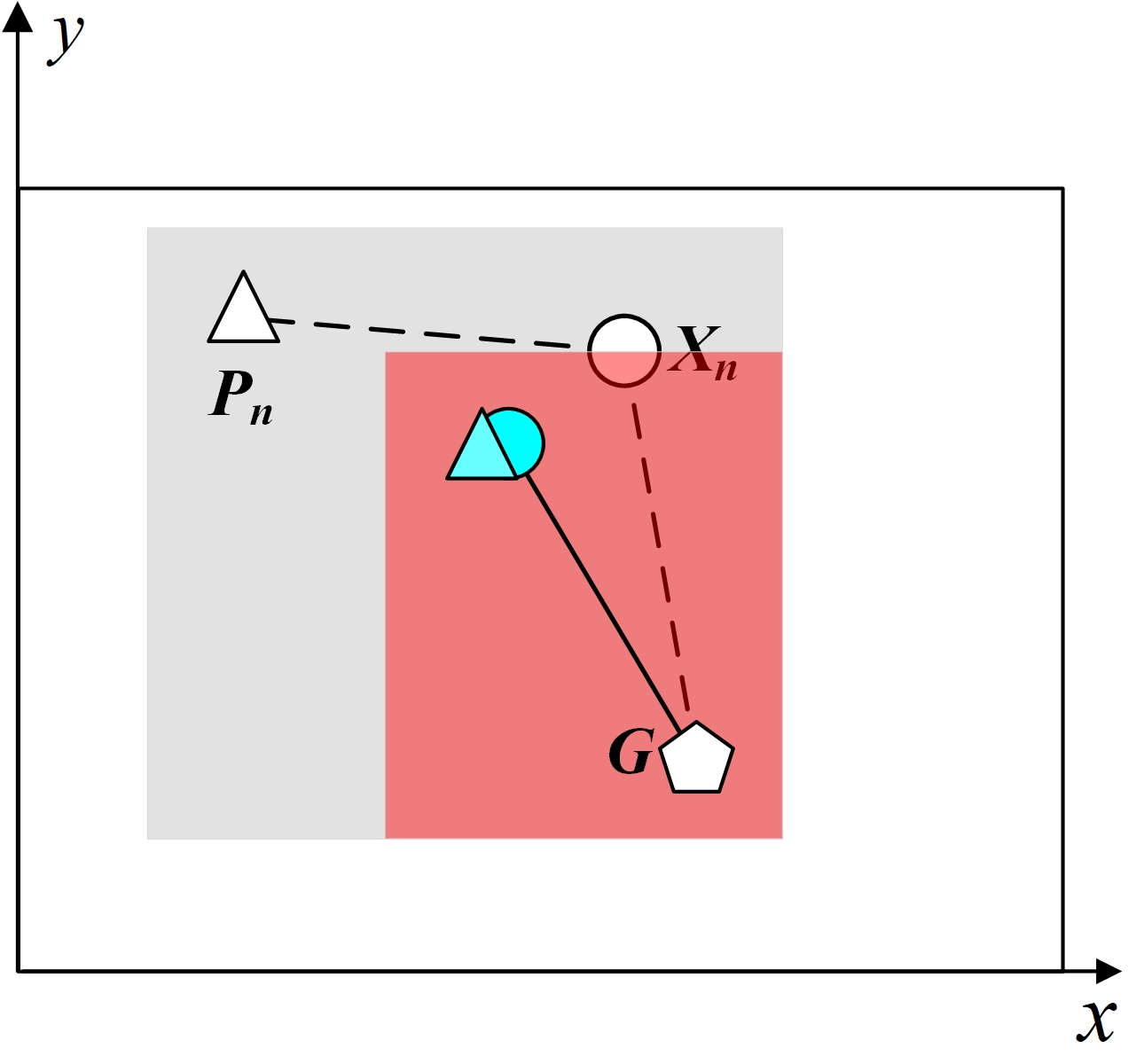}\\
		}
		\hspace{-2.5mm}
		\subfigure[]{		
			\includegraphics[width=1.4in]{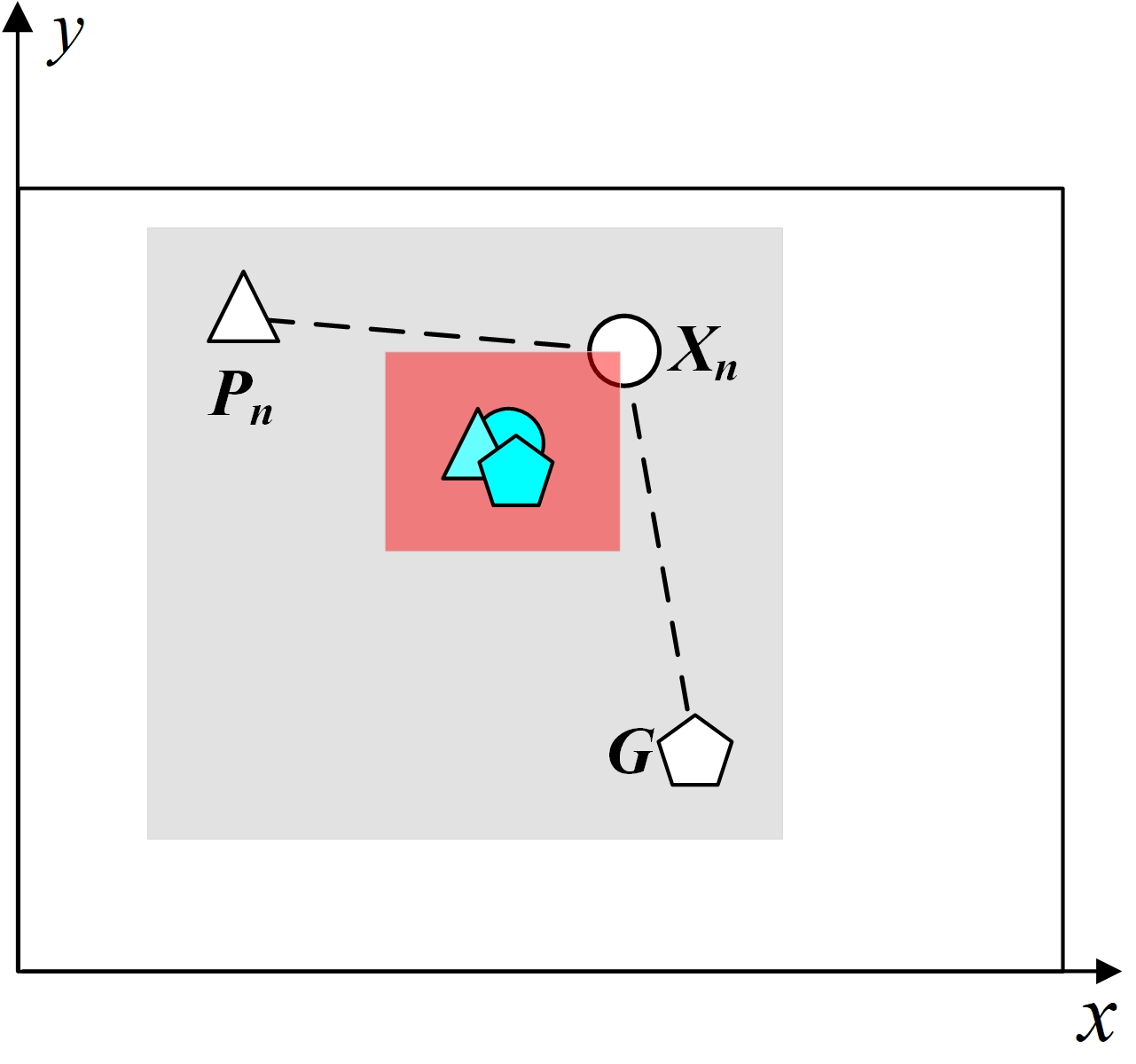}\\
		}%
		\hspace{-2.5mm}
		\subfigure[]{
			\includegraphics[width=1.4in]{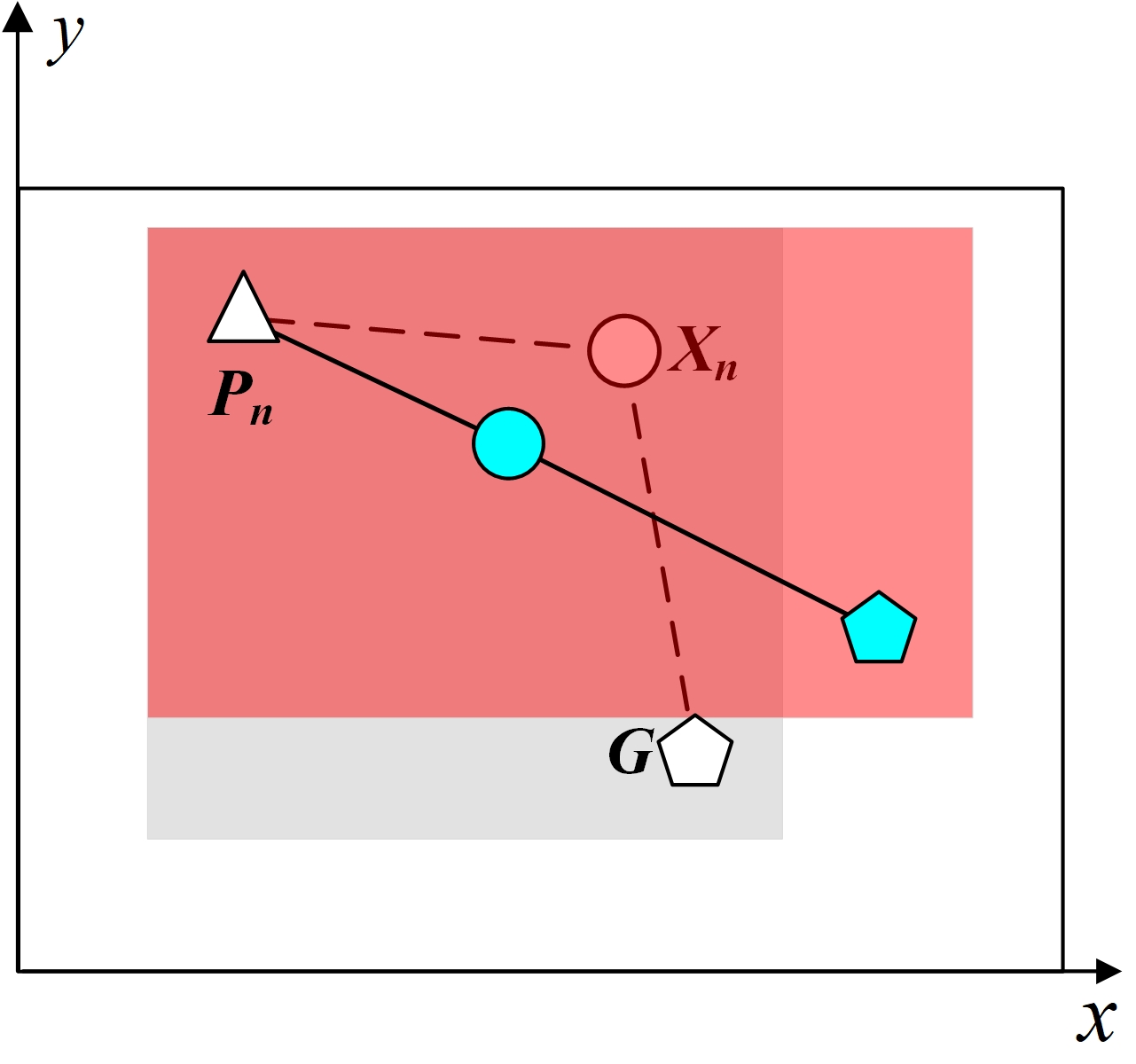}\\
		}
		\hspace{-2.5mm}
		\subfigure[]{
			\includegraphics[width=1.4in]{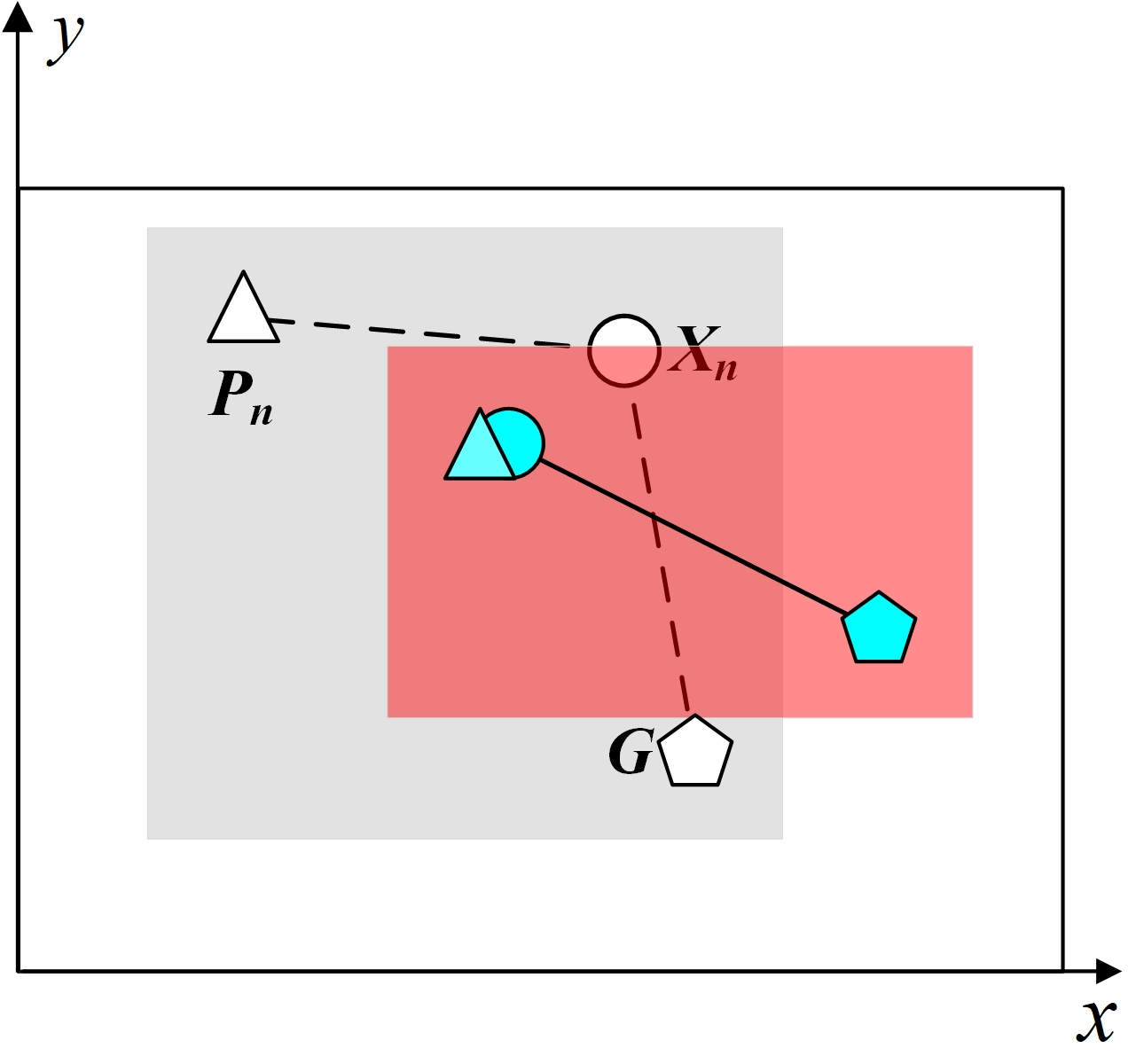}\\
		} 	
		\caption{\textit{n}th particle's potential search regions in two-dimensional space at $k$th and$(k+1)$th iterations.  (a) No $\boldsymbol{P}_n$ and $\boldsymbol{G}$ updated at $k$th iteration. (b) Only $\boldsymbol{P}_n$ updated  at $k$th iteration. (c) $\boldsymbol{G}$ updated by itself at $k$th iteration. (d)  No $\boldsymbol{P}_n$ updated but $\boldsymbol{G}$ updated by others at $k$th iteration. (e)  $\boldsymbol{P}_n$ updated by itself and  $\boldsymbol{G}$ updated by others at $k$th iteration.}
		\label{fig1}
	\end{figure*}

	The parameter tuning method garners considerable attention in balancing \textit{Er} and \textit{Ei}, owning to its straightforward implementation. This method primarily revolves aroud the adjustment of key parameters, namely inertia weight,  cognitive acceleration coefficient and  social acceleration coefficient. The core idea underlying this methodology is establishing a correlation between feedback information derived from the evolutionary environment and the balance of \textit{Er} and \textit{Ei}. 
	
   In the pursuit of achieving an optimal balance between the \textit{Er} and \textit{Ei}, researchers have diligently explored the correlation between feedback information (such as the number of iterations) and the associated parameters. This has boosted the development of a range of time-tuned-based parameters. Notable examples include  linear-decreasing inertia weight  proposed by Shi and Eberhart  \cite{shi1999empirical} and    time-varying acceleration coefficients introduced by Ratnaweera \textit{et al.} \cite{ratnaweera2004self}. Furthermore, scholars have delved into non-time feedback information, which has proven to be instrumental in achieving the desired balance between \textit{Er} and \textit{Ei}. For instance, the correlation between the involved parameters and particle-related attributes such as position \cite{nickabadi2011novel}, performance \cite{taherkhani2016novel}, and inter-particle distances \cite{wei2020multiple} is  investigated, which  yields valuable researches.

   Undoubtedly,  the establishment of coherent mathematical relationships between diverse  feedback information and parameters serves as a straightforward and efficacious means of attaining a harmonious balance between \textit{Er} and \textit{Ei}. Nevertheless,  the performance of PSO solely relying on parameter tuning method is inherently constrained by the absence of considerations pertaining to \textbf{\textit{P}}, \textbf{\textit{G}}, or population topology. Consequently, such method is frequently amalgamated with others to co-balance \textit{Er} and \textit{Ei} \cite{moazen2023pso}, \cite{chen2012particle}, \cite{ zhan2009adaptive}. In the context of this study, the integration of linear-decreasing inertia weight and time-varying acceleration coefficients into the proposed algorithm is  naturally executed, taking into account the influence of the number of function evaluations on the parameters.

	\begin{list}{}{}
	\item{ \emph{2) population topology construction}}
\end{list}

Population topology construction represents another widely applied method for balancing \textit{Er} and \textit{Ei}. Diverging from the parameter tuning, the core of this method lies in constructing the topology of current population (or multiple sub-populations) through the selection of distinct neighbors, and in this way controlling the \textit{Er} and \textit{Ei} abilities of the population (or multiple sub-populations), thereby  facilitating the balance between \textit{Er} and \textit{Ei}. 

 Kennedy \textit{et al}. \cite{kennedy2002population} theorized that different topologies have varying effects on balancing \textit{Er} and \textit{Ei}, and introduced five fixed  topological structures: all, ring, clusters, pyramid, and von Neumann. However,  the information of  particles involved in above structures is not fully used,  which prompts Mendes \textit{et al}. \cite{mendes2004fully}  to develop FIPS to make  individuals “fully informed”. Despite this advancement, the fixed structures struggle to effectively balance \textit{Er} and \textit{Ei} for more complex problems.  In response, researchers have devised numerous dynamic population topologies \cite{ zhang2011scale, zhang2021promotive} to enhance the balance of   \textit{Er} and \textit{Ei}. These dynamic topologies are often constructed using exemplars generated by diverse learning strategies, such as orthogonal learning strategy \cite{zhan2009orthogonal}, comprehensive learning strategy \cite{liang2006comprehensive}, and historical learning strategy \cite{li2015composite}, allowing for better adaptation to various complex problems. Moreover, employing multi-swarm \cite{van2004cooperative}, \cite{ li2015competitive}, \cite{niu2007mcpso} in a sense  can also be considered a form of population topology construction, offering additional way for  balancing \textit{Er} and \textit{Ei}.
 
Population topology construction has significantly enhanced the performance and generalization of PSO, which is mainly attributed to  implicitly controlling the guidance behavior of \textit{\textbf{P}}, \textit{\textbf{G}} for the population during the evolutionary process.  It is this implicit control that  creates opportunities for enhance the balance of \textit{Er} and \textit{Ei}. In light of this,  we, in this paper, 
describe the two bahaviors of \textbf{\textit{P}} and \textit{\textbf{G}} in standard PSO, and unveil the mechanisms underlying the two hebaviors, regarded as the research motivation for better balancing \textit{Er} and \textit{Ei} of PSO.
	
	\begin{list}{}{}
		\item{ \emph{3) Hybridization}}
	\end{list}
	
	 Hybridization,  involves incorporating the advantages of  other methods or algorithms  into PSO to address the challenge of balancing \textit{Er} and \textit{Ei}, although it introduces complexity. 
	
	Researchers have incorporated local search methods, such as Broyden–Fletch–Goldfarb–Shanno (BFGS) \cite{li2011hybrid}, \cite{cao2018comprehensive}, Nelder–Mead method \cite{cao2018comprehensive, fan2004hybrid},  into PSO, aiming to improve exploitation ability and thereby achieve a better balance between \textit{Er} and \textit{Ei}. Additionally, the hybridization of global search methods like  Levy Flight method \cite{hakli2014novel}, GA and Symbiotic Organisms Search \cite{farnad2018new} yields  better exploration ability, thereby also better balancing \textit{Er} and \textit{Ei}. Moreover, combining various particle swarm algorithms can also be considered as a form of hybridization. For instance, Lynn \textit{et al}. \cite{lynn2017ensemble} combines the strengths of  inertia weight PSO, HPSO-TVAC, FDR-PSO\cite{peram2003fitness}, LIPS and CLPSO algorithms to better balance \textit{Er} and \textit{Ei} of PSO.
	
 While hybridization does offer an improved balance between \textit{Er} and \textit{Ei}, it also introduces a more complex iterative optimization process, posing significant challenges in analyzing mathematical properties. To avoid this, we, in this paper, propose an algorithm that is consistent with the complexity of  standard PSO.

	\section{ Two behaviors of \textbf{\textit{P}} and \textbf{\textit{G}} in Standard PSO}\label{Double-edged sword features of PSO}
	
	These fruitful investigations have effectively improved the performance and  generalizability of PSO. Nonetheless, few scholars  have  paid attention to how \textbf{\textit{P}} and \textbf{\textit{G}} directly influence the balance of \textit{Er} and \textit{Ei}. Consequently, in this section,  we identify two specific behaviors:  \textbf{\textit{P}}'s uncontrollable-exploitation  and involuntary-exploration guidance behavior, and \textit{\textbf{G}}'s full-time and global guidance behavior, and delve into the underlying reasons causing the two behaviors. Moreover, we also elucidate the intricate mechanism by which the two bahaviors affect the balance between \textit{Er} and \textit{Ei}. 
	
	For understanding  the two behaviors of \textit{\textbf{P}} and \textit{\textbf{G}} in standard PSO, the values of $w$, $c_1$,  and $c_2$ are irrelevant factors.   Thereby, we let them be equal to one.  The previous velocity $v_n^d$ has an impact on  the location of the potential search region, which is often confined within the range $[-\underset{max}{v^d}, \underset{max}{v^d}]$, where $\underset{max}{v^d}$ represents the maximum  velocity limit on $d$ dimension. To more effectively illustrate the influence of the previous velocity, we extend $v_n^d$ to the interval $\mathscr{V}^d_n = [-\underset{max}{v^d}, \underset{max}{v^d}]$. As a consequence,  a simple updating fomula is  formed as shown in (\ref{simplified condition pso equat 1}). Based on this, the potential search region of one particle in PSO is described  within a two-dimensional space, as illustrated in Fig \ref{fig1}, wihch aids in understanding the two behaviors of \textbf{\textit{P}} and \textit{\textbf{G}} and the underlying  mechanisms by which they affect the balance between \textit{Er} and \textit{Ei}.
	
		\begin{equation} \label{simplified condition pso equat 1}
		{v}_n\leftarrow  \mathscr{V}^d_n + {r}_1  \cdot 	({p}^d_n-{x}^d_n)+{r}_2 \cdot ({g^d}-x^d_n)
	\end{equation}

	\subsection{ \textbf{\textit{P}}'s Uncontrollable-Exploitation  and Involuntary-Exploration Guidance Behavior}\label{1 Phenomenon}
	
	Fig. \ref{fig1} shows the potential search regions of $n$th particle at the $k$th and$(k+1)$th iterations under different update statuses of $\boldsymbol{P}_n$ and \textit{\textbf{G}}. The potential search regions of $k$th and $(k+1)$th iterations are marked with a shade of gray and red, respectively. As shown in Fig. \ref{fig1} (a)-(c), in the absence of the updated $\boldsymbol{G}$ by other particles during the $k$th iteration, $\boldsymbol{P}_n$, under the influence of \textit{\textbf{G}}, continually guides the particle to exploit around the previous search region regardless of the quality of the previous search region. We characterize this phenomenon  as $\boldsymbol{P}$'s uncontrollable-exploitation guidance behavior, which may lead to the problem that one particle exploits less time in the superior region and more time in inferior one, thereby weakening the efficient utilization of computational resources for exploitation and  disrupting the delicate balance between \textit{Er} and \textit{Ei}. Moreover, as observed in Fig. \ref{fig1}(d)-\ref{fig1}(e), when \textbf{\textit{G}} is updated by other particles, $\boldsymbol{P}_n$, under the influence of \textit{\textbf{G}}, passively drives the particle to slowly jump out of the previous search region. we characterize $\boldsymbol{P}$ as exhibiting involuntary-exploration guidance behavior, which hinders particles from voluntarily exploring other unknown regions of search space devoid of \textbf{\textit{G}}'s influence, weakening exploration ability and further negatively affecting on the balance between \textit{Er} and \textit{Ei}.
	
	Analysing further, it is found that the existence of \textbf{\textit{P}}'s behavior stems from a degree of being subjugation to the influence of \textbf{\textit{G}}. Besides, \textit{\textbf{P}}'s uncontrollable-exploration guidance behavior emerges due to the uncontrollable number of exploration  attributed to the absence of reasonable assessment to current search region's quality, \textit{\textbf{P}}'s involuntary-exploration guidance behavior is also rooted in its exclusive reliance on information from its corresponding particle.

	\subsection{\textit{\textbf{G}}'s Full-Time and Global Guidance Behavior}
In standard PSO, on the one hand, \textit{\textbf{G}} always guides particles to move toward the region where \textit{\textbf{G}} is located throughout the iteration process, no matter whether the quality of \textit{\textbf{G}} is good or bad \cite{zhang2021promotive}. We call this behavior as the full-time guidance behavior \textit{\textbf{G}}. Due to the higher possible poor quality of \textit{\textbf{G}} (especially in multimodal functions) in pre-iteration period, this behavior may cause that one particle tends to be pulled into an inferior region to explore, wasting the exploration resource, thereby negatively impacting the balance of \textit{Er} and \textit{Ei}. On the other hand, on each iteration slice, all particles are guided by \textit{\textbf{G}} to cluster towards the local region where \textit{\textbf{G}} is located \cite{lynn2015heterogeneous}. This behavior of \textbf{\textit{G}} is characterized as global guidance behavior, which may lead to the problem that all particles are guided by \textbf{\textit{G}} and cluster towards the local region in case of \textbf{\textit{G}} at the local optimum during each iteration slice, causing a rapid loss of particle diversity (especially on multimodal problems), weakening the exploration ability, and further negatively affecting the balance between \textit{Er} and \textit{Ei}.

Moreover, according to the features of solo iterative channel and single population in PSO, it is clear that the existence of \textit{\textbf{G}}'s behavior emanates from each particle's absolute reliance on single updating channel and the absence of control mechanism of its behavior.  It is  also worth noting that \textit{\textbf{P}} and \textit{\textbf{G}} are coupled together to control one particle's state under the influence of these features, making it more difficult to achieve balance between \textit{Er} and \textit{Ei}.

 Based on the analysis above, aiming to balance \textit{Er} and \textit{Ei} of PSO, the following challenges must be addressed: 1) How to mitigate the influence of \textbf{\textit{G}} on \textit{\textbf{P}} and eliminate the status quo of each particle's absolute reliance on \textit{\textbf{G}}, so as to decouple the \textit{\textbf{P}} and \textbf{\textit{G}} as much as possible. 2) How to dispel \textit{\textbf{P}}'s sole reliance on information about its corresponding particle to empower \textit{\textbf{P}} with voluntary-exploration guidance behavior. 3) How to reasonably assess the quality of the search region and control the number of search region searches, so as to control \textit{\textbf{P}}'s exploration guidance behavior. 4) How to establish the control mechanism of \textit{\textbf{G}}, so as to control \textbf{\textit{G}}'s behavior.

\section{Methods}\label{DCPSO-ABS}
\begin{figure}[!t]
	\centering
	\includegraphics[width=3.5in]{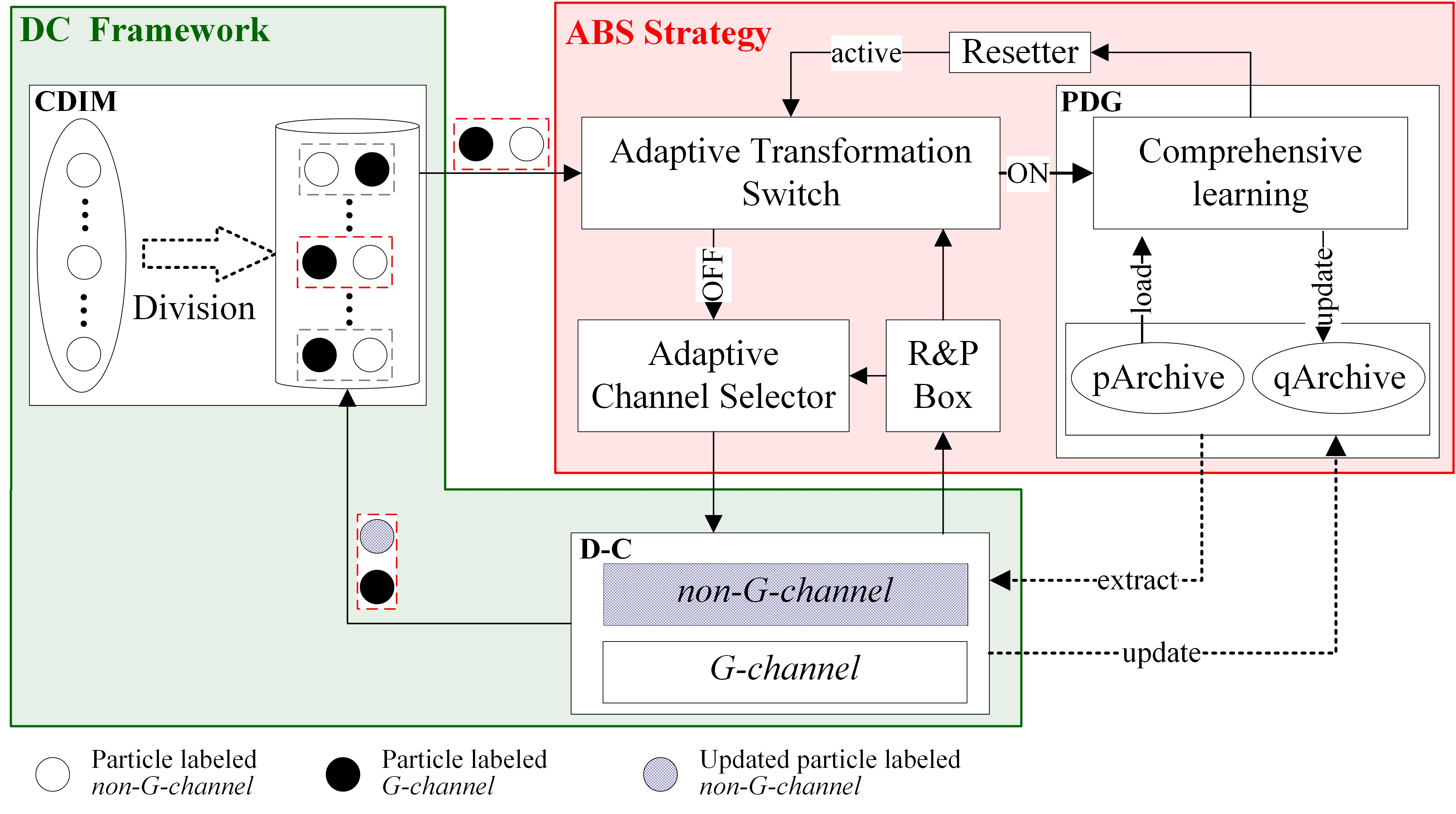}
	\caption{Architecture of DCPSO-ABS algorithm.}
	\label{Architecture of DCPSO-ABS}
\end{figure}	
	Aiming to solve the above challenges for  balancing \textit{Er} and \textit{Ei}, a DCPSO-ABS is  propounded. The architectural schematic of DCPSO-ABS is delineated in Fig. \ref{Architecture of DCPSO-ABS}. The proposed algorithm encompasses a  DC framework and an ABS strategy. The DC framework, comprising dual-channeller (D-C) and cell division based initialization method (CDIM) is introduced to 
	mitigate the influence of \textbf{\textit{G}} on \textit{\textbf{P}} and break the status quo of each particle's absolute reliance on the unique update channel in standard PSO, for conveniently controlling of the two bahaviors of \textit{\textbf{P}} and \textbf{\textit{G}}.  The ABS strategy incorporates pivotal elements: promising direction generator  (PDG), adaptive transformation switch, adaptive channel selector, reward and penalty box (R $\&$ P Box), and resetter, for empowering  \textbf{\textit{P}} with voluntary-exploration guidance behavior, adaptive controlling \textbf{\textit{P}}'s  exploitation guidance behavior as well as \textit{\textbf{G}}'s full-time and global guidance behavior.   These critical components are explicated in detail below.

	\subsection{DC Framework}\label{Initialization}
	As previously discussed, mitigating the influence of \textbf{\textit{G}} on \textit{\textbf{P}}, and breaking the status quo of each particle's absolute reliance on the \textit{\textbf{G}} is essential to regulate the two bahaviors of  \textit{\textbf{P}} and  \textit{\textbf{G}}. In response, a DC framework including  D-C and CDIM is constructed to  ensure that \textit{\textbf{P}} is not entirely subject to the effects of \textbf{\textit{G}} and  that not all particles are influenced by \textit{\textbf{G}}, respectively.  This framework serves as a prerequisite to effectively  manage the behaviors of \textit{\textbf{P}} and \textit{\textbf{G}}.
	
		\textit{1) D-C:}\label{ Dual-Channeler} To mitigate the influence of \textbf{\textit{G}} on \textit{\textbf{P}} as much as possible, a D-C,  including  \textit{non-G-channel} and \textit{G-channel}, is constructed.  The formulas for \textit{non-G-channel}  and \textit{G-channel} are delineated as follows:
				
	\begin{equation} \label{non-G}
		\textit{non-G-channel}: 	 
		\begin{cases}
			\underset{non-G}{{v}_n^d} = w\cdot \underset{non-G}{{v}_n^d}+{c}\cdot {r} \cdot ({p}_n^d-\underset{non-G}{{x}_n^d})\\
			\underset{non-G}{{x}_n^d} = \underset{non-G}{{x}_n^d} + \underset{non-G}{{v}_n^d}
		\end{cases}
	\end{equation}
	and
	\begin{equation} \label{G}
		\textit{G-channel}:
		\begin{cases}
			\underset{G}{{v}_n^d} = w\cdot \underset{G}{{v}_n^d}+{c_1}\cdot {r_1} \cdot ({p}_n^d-\underset{G}{{x}_n^d})\\+{c_2}\cdot {r_2} \cdot ({g}^d-\underset{G}{{x}_n^d})\\
			\underset{G}{{x}_n^d}=\underset{G}{{x}_n^d}+\underset{G}{{v}_n^d}
		\end{cases}.
	\end{equation}

Indeed, the distinct operational principle of the \textit{non-G-channel} being solely guided by \textbf{\textit{P}} and the \textit{G-channel} retaining the guidance from \textbf{\textit{G}} aids in mitigating the influence of \textbf{\textit{G}} on \textit{\textbf{P}} within the D-C framework. This operation allows for a certain degree of independence and autonomy in the guidance mechanism, which makes the management of the behaviors of \textit{\textbf{P}} and \textbf{\textit{G}} more effectively.

	\textit{2) CDIM:}\label{ Cell Division-based Initialization Method}	
With the objective of breaking the status quo of each particle's absolute reliance on \textit{\textbf{G}} in standard PSO,  a CDIM is introduced. In CDIM, an initial set of $N$ particles is randomly generated by the  traditional PSO initialization method. Each particle undergoes a process akin to cellular division \cite{scholey2003cell}, drawing the creation of two particles with identical information.  These pairs of particles, sharing the same information, are then grouped together as one sub-swarm, and are randomly labeled as either \textit{non-G} or \textit{G} to indicate the specific channel they serve throughout their lifecycle. This reliance mode ensures not all particles are absolute reliance on \textit{\textbf{G}}, thereby further contributing to the regulation of bahaviors of \textit{\textbf{P}} and \textit{\textbf{G}}.

	\subsection{ABS Strategy }\label{Adaptive balance Search Strategy} 
	In this section, the ABS strategy is introduced to empower $\boldsymbol{P}$  with voluntary-exploration behavior, adaptive control \textbf{\textit{P}}'s  exploitation behavior as well as \textit{\textbf{G}}'s full-time and global behavior. The ABS strategy unfolds through several key components: 1) PDG, which functions to  empower $\boldsymbol{P}$ with voluntary exploration guidance behavior. 2) Adaptive channel selector, which is constructed to mainly adaptive control of \textit{\textbf{G}}'s full-time and global guidance behavior. 3) R \& P box,  which plays a determinative role in empowering \textbf{\textit{P}} with controllable-exploitation guidance behavior. 4) Adaptive transformation switch, which serves the purpose of determining whether \textit{\textbf{P}}'s exploitation guidance behavior should transition into exploration guidance behavior. 5) Resetter, which ensures the harmonious co-operation of the components among the various constituents comprising the ABS strategy.

			\textit{1) PDG:} As previously discussed, the key of empowering $\boldsymbol{P}$  with voluntary-exploration behavior is amalgamation of other particles' information, thereby introducing new direction elements to current \textbf{\textit{P}}.  With regards to this, PDG, including   Personal archive (\textit{pArchive}) and \textit{\textbf{G}}, Comprehensive learning (CL) strategy and \textit{qArchive}, is proposed to generate a new $\boldsymbol{Q}_n$ to replace $\boldsymbol{P}_n$ for $n$th sub-swarm, so as to guide the consponding subswarm  to explore other unknown regions of search space.

		\textit{\textit{pArchive} and \textit{\textbf{G}}:} In DCPSO-ABS, each sub-swarm is restricted to have only one  $\boldsymbol{P}$.  \textit{pArchive}, in our proposed algorithm, consists of updated $\boldsymbol{P_n}$ from each sub-swarm. By comparing all $\boldsymbol{P_n}$, the best one is selected to searve as \textbf{\textit{G}} for the whole population. The updated formula of $\boldsymbol{P_n}$  is shown in (\ref{pbest update formular}). Besides,  the \textit{pArchive} is used to add new direction elements to current \textbf{\textit{P}}.

 \textit{CL Strategy and \textit{qArchive}:} The strategic infusion of other direction elements into  extant  \textbf{\textit{P}}  based on \textit{pArchive} is a significant inquiry. Among the learning strategies, CL strategy exhibits strong global exploration ability due to allowing one particle to potentially be guided by all other particles \cite{liang2006comprehensive}. Therefore, this unique feature has the potential to  empower \textit{\textbf{P}} with voluntary-exploration behavior. Yet as the number of individuals increases, CL strategy may suffer from weak convergence due to  the excessive learning exemplars. To alleviate this deficiency, in our proposed algorithm, we treat the sub-swarms as individuals, and modify the individual-to-all individuals learning strategy of CL strategy to a sub-swarm-to-all sub-swarms learning ones, halving the number of learning exemplars.  And then, we generate $\boldsymbol{Q}_n$ of $n$th sub-swarm  from $\boldsymbol{P}_n$s of all sub-swarms (not all individuals) based on Fig. 2 in \cite{liang2006comprehensive}. All $\boldsymbol{Q_n}$s collectively form \textit{qArchive}. 

\begin{figure}[!t]
	\centering
	\includegraphics[width=2.5in]{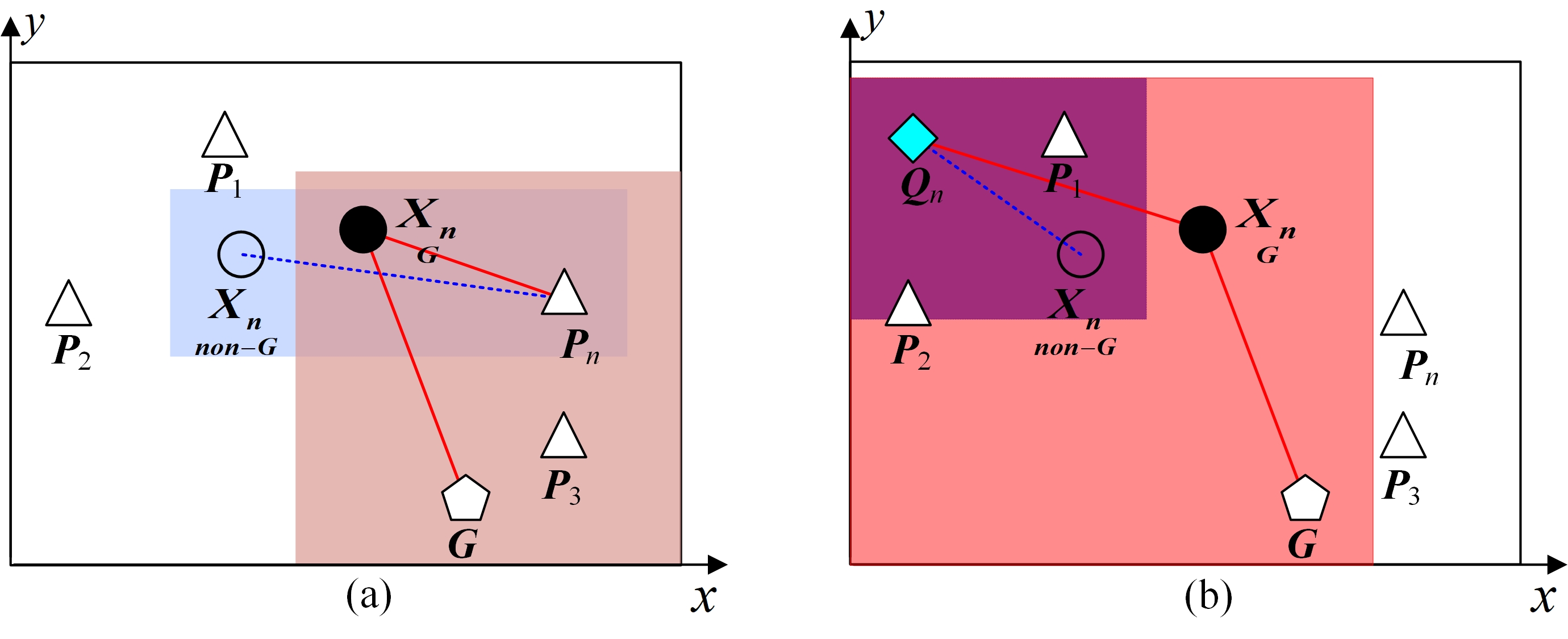}
	\caption{Potential search regions under \textit{non-G-channel} (blue) and \textit{G-channel} (red). (a) Influenced by \textbf{\textit{P}}. (b) Influenced by \textit{\textbf{Q}}.}
	\label{Two kinds of search regions in two-dimensional space}
\end{figure}

%

		\newcounter{TempEqCnt}
\setcounter{TempEqCnt}{5}
\setcounter{equation}{4}

\begin{figure*}[b]
	\hrulefill
	\begin{align}
		\boldsymbol{P_n}(k+1)=
		\begin{gathered} 
			\left\{ \begin{gathered}
				\underset{non-G}{\boldsymbol{X_n}}(k+1)*(fit(	\underset{non-G}{\boldsymbol{X_n}}(k+1)) \leq fit(\boldsymbol{P_n}(k)))+\boldsymbol{P_n}(k)*(fit(\underset{non-G}{\boldsymbol{X_n}}(k+1))>fit(\boldsymbol{P_n}(k)))\\
				$\textit{or}$\\				
				\underset{G}{\boldsymbol{X_n}}(k+1)*(fit(\underset{G}{\boldsymbol{X_n}}(k+1))\leq fit(\boldsymbol{P_n}(k)))+\boldsymbol{P_n}(k)*(fit(\underset{G}{\boldsymbol{X_n}}(k+1))>fit(\boldsymbol{P_n}(k)))\\
			\end{gathered}  \right. \hfill \\
		\end{gathered}
		\label{pbest update formular}
	\end{align}
\end{figure*}
\setcounter{equation}{\value{TempEqCnt}}

By replacing \textbf{\textit{P}} with \textbf{\textit{Q}}, the proposed algorithm  gains the ability  to empower \textit{\textbf{P}} with voluntary-exploration guidance behavior. As shown in Fig. \ref{Two kinds of search regions in two-dimensional space}, whether in \textit{non-G-channel} or  \textit{G-channel}, the utilization of \textit{\textbf{Q}} instead of \textit{\textbf{P}} brings a rapid transformation of the potential search regions, enabling  \textit{\textbf{P}} to voluntarily guide the particles to explore  unknown regions of the search space without the influence of \textit{\textbf{G}}, which is aligning with the concept of voluntary-exploration guidance behavior. 

			\begin{figure}[t]
	\centering
	\includegraphics[width=2.5in]{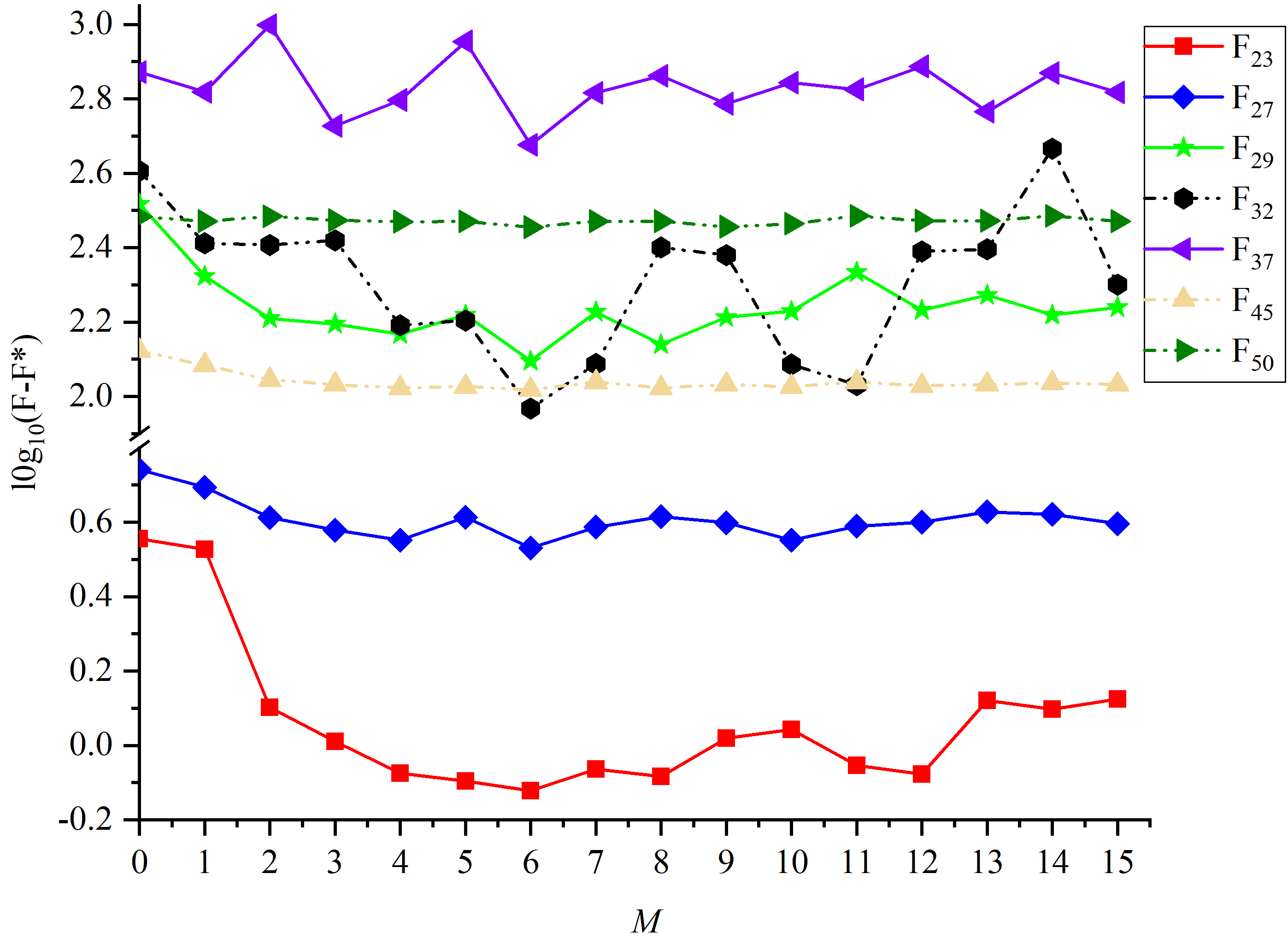}
	\caption{Performance of DCPSO-ABS  with various values of $M$.}
	\label{M}
\end{figure}

        \textit{2) Adaptive Channel Selector:}\label{ Adaptive Channel Selector} As elucidated earlier, establishing a reasonable mathematical relation between the ratio of particles guided by \textit{\textbf{G}} to those not guided by \textbf{\textit{G}}, and the number of iterations is a pivotal point for adaptive controlling of \textit{\textbf{G}}'s full-time and global guidance behavior. In response to this imperative, based on the DC framework, an adaptive channel selector is designed to discern the optimal channel (\textit{non-G-channel} or \textit{G-channel}) for engagement during each iteration.
        This strategic choice dictates whether  \textbf{\textit{G}} should be involved in the process, thereby exerting control over its full-time and global guidance behavior. This selection process is intricately governed by three factors: refreshing gap, adaptive function, and channel select condition, detailed as follows.
   
\textit{	Refreshing Gap (\textit{M}):} The refreshing gap is introduced to determine an upper limit on the number of times that one promising direction is used, and thus determine the upper limit on the number of times that each sub-swarm exploits in promising regions. As the impact of different values of 
	$M$ on the algorithm's performance varies, fine-tuning of the refreshing gap becomes essential \cite{liang2006comprehensive}. This section delves into empirical  studies on relevant functions, such as Shifted and Rotated Rosenbrock’s function, Shifted and Rotated Non-Continuous Rastrigin’s function, Shifted and Rotated Schwefel’s function, Hybrid function 3 (N=3), Hybrid function 6 (N=5), Composition function 3 (N=4) and Composition function 7 (N=6) listed in Table S-I in the Supplementary Material. The influence of values of \textit{M} ranging from 0 to 15 on each function in 10-D is investigated.   Fig.\ref{M} presents the  empirical results, indicating that a value of $M$ approximate 6 indeed confers notable benefits for the DCPSO-ABS algorithm. Consequently, in our experiments, the refreshing gap $M$ is uniformly fixed at 6 for all experiments.
	
	\textit{Adaptive Function:} The adaptive function regarding to the number of function evaluations is devised. It works for establishing an upper threshold dictating the upper limit number that \textit{non-G-channel} and \textit{G-channel} are involved in one determined promising direction.  The adaptive function is formulated as

		\setcounter{equation}{5}
		\begin{equation} \label{dynamic adjustment function m1}
		\begin{cases}
		M_1=\lceil M \cdot(1-\dfrac{FEs}{FEs_{max}}) \rceil \\
			M_2=M-M_1
		\end{cases}
	\end{equation}	where $M_1$ and $M_2$ denote the upper limits on the number that 	\textit{non-G-channel} and 	\textit{G-channel} are used in one determined promising direction within $M$, $FEs_{max}$ denotes the maximum function evaluations, $FEs$ denotes the  function evaluations counter,  $\lceil\cdot\rceil$ represents rounding down. Evidently,  at a constant value of $M$, the proportion of $M_1$ to  $M_2$ is decreasing with as $FEs$ increases. This trend ensures preservation of population diversity in pre-iteration period and the convergence in later iteration period, which implies the adaptive function has a positive effect on controlling the behavior of \textit{\textbf{G}}.
	
	 \textit{Channel Select Condition:} Since \textit{G-channel} is capable of rapidly aggregating particles, it is necessary to design the serial relationship in which the \textit{non-G channel} operates before the \textit{G channel}. In this paper,  \textit{non-G-channel} is activated under the condition
	 
	\begin{itemize}
		\item   $\alpha_n \leq M_1$
	\end{itemize}
	where $\alpha_n$ denotes the number of times that the particle serving \textit{non-G-channel} in the $n$th sub-swarm has already searched in one determined promising region.
	
	The \textit{G-channel}  is activated only if the following condition meets
	\begin{itemize}
		\item $ M_1 < \alpha_n \leq M     \wedge  \beta_n=0$
	\end{itemize}
	where $\beta_n$ represents whether $\boldsymbol{P}_n$ is  updated in 	\textit{non-G-channel}.  $\beta_n=0$ means $\boldsymbol{P}_n$ is not updated in 	\textit{non-G-channel}. On contrast,  $\beta_n=1$ means $\boldsymbol{P}_n$ is  updated in 	\textit{non-G-channel}.
	
	Clearly, influenced by the joint action of refreshing gap, adaptive function and channel select condition,  the adaptive channel selector  assumes a pivotal role in adaptively controlling \textit{\textbf{G}}'s full-time and global guidance behavior.

		\textit{3) R $\&$ P Box:}	R \& P box is constructed to determine whether to reward or penalize the number of searches in promising regions according to particle's update state, so as to adaptively control \textbf{\textit{P}}'s exploitation guidance behavior. For \textit{non-G-channel}, if $\boldsymbol{P}_n$ is updated, reset $\alpha_n = 0$, rewarding an additional number of searches. Otherwise, $\alpha_n = \alpha_n + 1$, penalizing one less search in the current promising region. For \textit{G-channel}, if $\boldsymbol{G}$ is updated, reset $\alpha_n = M_1+1$, rewarding an additional number of searches. If $\boldsymbol{P}_n$ is updated, but $\boldsymbol{G}$ is not, then $\alpha_n = \alpha_n $, rewarding one more search in the current promising region. If $\boldsymbol{P}_n$ is not updated, then $\alpha_n = \alpha_n + 1$, penalizing one less search in the currently promising region. By establishing the above rules for rewards and penalties, the R \& P box plays a determinative role in controlling \textbf{\textit{P}}'s exploitation guidance behavior.

	\textit{4)  Adaptive Transformation Switch:} In conjunction with the aforementioned methodologies to regulate the behaviors of \textit{\textbf{P}} and \textbf{\textit{G}}, careful consideration must be given to the timing of \textbf{\textit{P}}'s reconstruction by PDG. Therefore,	in this section, an adaptive transformation switch is constructed to decide whether one given sub-swarm  enters PDG to generate new promising direction or enters adaptive channel selector to  participate in the iterative process, thereby deciding the timing when the corresponding \textbf{\textit{P}} is reconstructed by PDG.
	
	We set that when the adaptive transformation switch satisfies the specified condition,  it is activated, prompting the sub-swarm to engage with the PDG. Conversely, if the condition is not met, the switch remains inactive, guiding the sub-swarm towards the adaptive channel selector for further processing. The specified condition is 
	
	\begin{itemize}
		\item  $(\alpha_n > M_1 \wedge \beta_n=1) \vee ( \alpha_n > M )$.
	\end{itemize}

 By constructing the specified condition, the adaptive transformation switch can be applied to decide the timing when P is reconstructed by PDG, essentially deciding whether \textit{\textbf{P}}'s exploitation guidance behavior should be turned exploration guidance behavior.

			\textit{5) Resetter:} \label{Resetter}
 In resetter, two parameters are reset: $\alpha_n =0 $ and $ \beta_n =0 $, ensuring the co-operation of components in ABS strategy.

	\begin{algorithm}[!t]
		\renewcommand{\algorithmicrequire}{\textbf{Input:}}
		\renewcommand{\algorithmicensure}{\textbf{Output:}}
		\caption{The pseudocode of ABS strategy}
		\label{strategy 2}
		\begin{algorithmic}[1]
			\Require  $\boldsymbol{P}_1, \boldsymbol{P}_2,...,\boldsymbol{P}_N $, $\alpha_n$, $\beta_n$, $M_1$ , $M$, $\underset{non-G}{\boldsymbol{X}_n}$, $\underset{non-G}{\boldsymbol{V}_n}$, $\underset{G}{\boldsymbol{X}_n}$, $\underset{G}{\boldsymbol{V}_n}$, $FEs$; 
			\Ensure  $\alpha_n$, $\beta_n$, $\boldsymbol{P}_n$, $\boldsymbol{G}$, $FEs$ ; 		
			
			\State Caculate $M_1$ and $M_2$ according to (\ref{dynamic adjustment function m1});
			\If{{$(\beta_n \neq 0 \&\& \alpha_n > M_1) \| (\beta_n =0 \&\& \alpha_n > M$})}
			\State $\beta_n =0$, $\alpha_n =0$, and generate $\boldsymbol{Q}_n$;
			\ElsIf {$\alpha_n  \leq  M_1$}
			\State Update $\underset{non-G}{\boldsymbol{V}_n}$ , $\underset{non-G}{\boldsymbol{X}_n}$ through (\ref{non-G}), $FEs=FEs+1$;
			
			\If {$f(\underset{non-G}{\boldsymbol{X}_n}) \geq f(\boldsymbol{P}_n)$}
			\State $\alpha_n=\alpha_n+1$;
			\Else
			\State $\alpha_n=0$, $\beta_n=1$, and update $\boldsymbol{P}_n$;
			\If{$f(\boldsymbol{P}_n) < f(\boldsymbol{G})$}
			\State Update $\boldsymbol{G}$;
			\EndIf
			\EndIf
			\ElsIf{$M_1 < \alpha_n \leq M \&\& \beta_n=0$}
			
			\State Update $\underset{G}{\boldsymbol{V}_n}$, $\underset{G}{\boldsymbol{X}_n}$ by (\ref{G}), $FEs=FEs+1$;
			\If {$f(\underset{G}{\boldsymbol{X}_n}) \geq f(\boldsymbol{P}_n)$}
			\State $\alpha_n=\alpha_n+1$;
			\Else                     
			\State  Update $\boldsymbol{P}_n$;
			\If{$f(\boldsymbol{P}_n) < f(\boldsymbol{G})$}
			\State $\alpha_n=M_1+1$, and update $\boldsymbol{G}$;
			\EndIf
			\EndIf			     
			\EndIf
			\State \Return  $\alpha_n$, $\beta_n$, $\boldsymbol{P}_n$, $\boldsymbol{G}$, $FEs$.
		\end{algorithmic}
	\end{algorithm}

				\begin{figure}[t]
		\centering
		\includegraphics[width=1.9in]{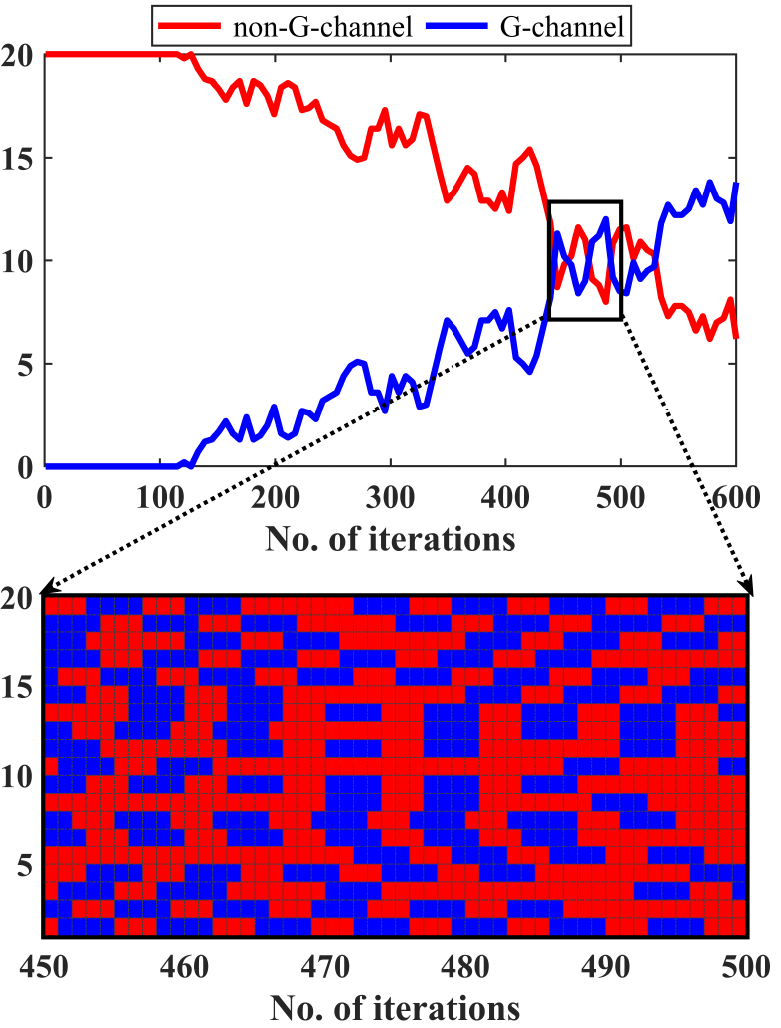}
		\caption{The number that each channel is employed changes with the number of iterations (up). The combination of two channels employed by the whole swarm on different iterations (down).}
		\label{Plot of changes in sub-swarms participation}
	\end{figure}

In conlusion, the combined operation of the five components enables ABS strategy to empower $\boldsymbol{P}$  with voluntary-exploration guidance behavior, adaptive control \textbf{\textit{P}}'s  exploitation guidance behavior, as well as \textit{\textbf{G}}'s full-time and global guidance behavior.	
The pseudocode for ABS strategy is presented in Algorithm  \ref{strategy 2}. Interestingly, the ABS strategy pulls a unique combination property. Specifically, a total of $N$ particles are involved in each iteration due to the sequential relationship between the two channels, even though there are a total of $2N$  particles available. Consequently, the maximum number of combinations of $N$ particles involved in each iteration is $2^N$, and correspondingly, the maximum number of combinations of channels to be employed in each iteration is also $2^N$.  The intuitive illustration of this unique combination is depicted in Fig. \ref{Plot of changes in sub-swarms participation}. The up-subplot illustrates the trend that the number of each channel employed varies with the number of iterations. This subplot shows that the number of  employed  \textit{non-G-channel} decreases while  the number of  employed  \textit{G-channel} increases with the progression of iterations.  Moreover, it also shows that \textit{G-channel} is not employed until the almost 100th iteration. These nuanced controls of \textit{non-G-channel} and \textit{G-channel} may ensure preservation of population diversity in pre-iteration period and the convergence in the later iteration period and thereby aid to balance \textit{Er} and \textit{Ei}.
The down-subplot shows that the varying combinations of  two channels employed at different iteration stages are different. This variance across iterations yields subtle oscillations of diversity, possibly fostering algorithmic activity, further positively affecting on the balance  between \textit{Er} and \textit{Ei}.

%
%
%

	\subsection{General Algorithm of DCPSO-ABS}
	
	From above, the DCPSO-ABS algorithm is proposed, whose pseudocode is presented in Algorithm  \ref{VRD-PSO}. To expound on the properties of the proposed algorithm, we delve into its  exploration and exploitation abilities, and theoretically derive the computational complexity of the algorithm.

\textit{1) Exploration and Exploitation Abilities:} The metric, swarm diversity, is widely used to identify whether the swarm  conducts exploration or exploitation \cite{olorunda2008measuring}. The diversity is commonly measured by

			\begin{equation} \label{diversity eqation 1}
		\begin{cases}
		Divesity(k)=\dfrac{\sum_{n=1}^{N} \sqrt{\sum_{d=1}^{D} \big(x_i^d(k)-\bar{x}^d(k)\big)^2}}{N}  \\
		\bar{x}^d(k)=\dfrac{\sum_{n=1}^{N} x_i^d(k)}{N}
		\end{cases}.
	\end{equation}

\begin{figure}[!t]
		
		\subfigure[]{
			
			\includegraphics[width=1.5in]{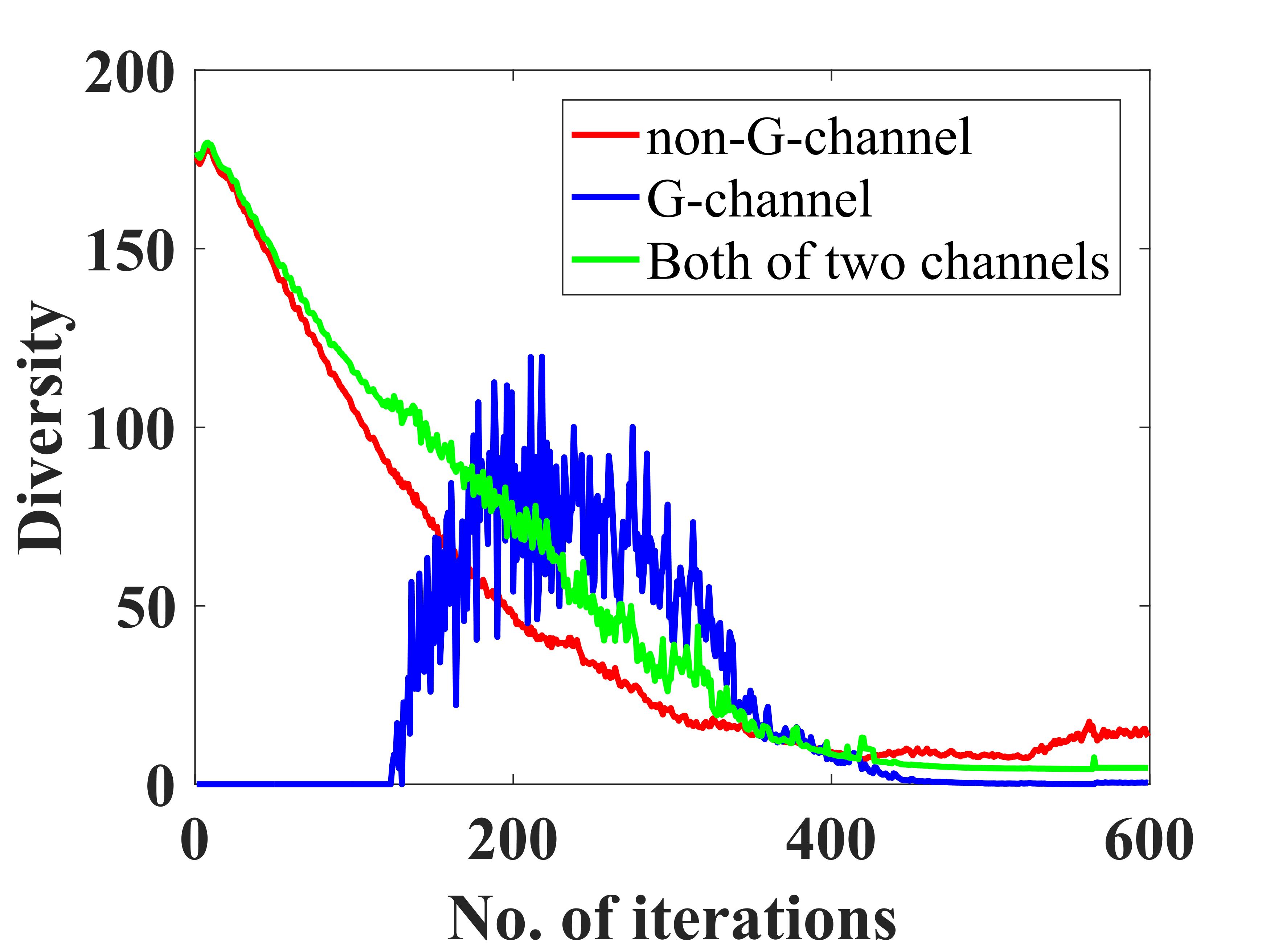}\\
		}%
		\hspace{2mm}
		\subfigure[]{
			\includegraphics[width=1.5in]{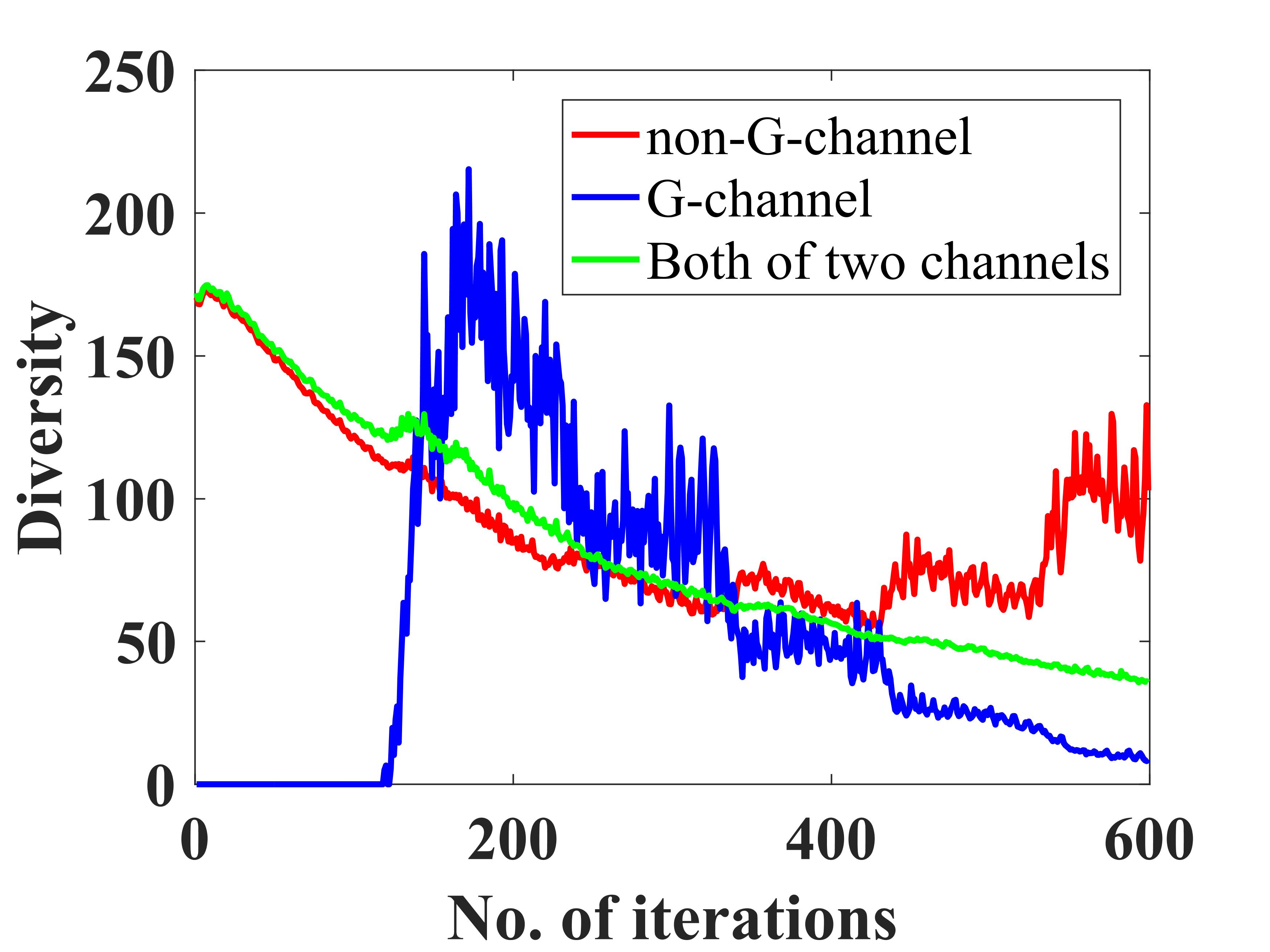}\\
		}
		\centering
		\caption{Comparison of Diversity influenced by   \textit{non-G-channel}, \textit{G-channel} and both of two channels.  (a) Sphere Function.   (b)  Rastrigin Function.}
		\label{Comparison of Diversity}
	\end{figure}

Assume there are $N_1$ and $N_2$ sub-swarms (where $N_1+N_2=N$) involved in \textit{non-G-channel} and \textit{G-channel} in each iteration, respectively. The diversities influenced by these channels individually and collectively are investigated for the Sphere Function and Rastrigin Function on 10-D, and corresponding results are illustrated in Fig. \ref{Comparison of Diversity}. In the early iteration stage (\textit{1}th-\textit{140}th), the algorithm relied on \textit{non-G-channel} exhibits high diversity, showing strong exploration ability. Subsequently, in the middle iteration stage (\textit{140}th-\textit{350}th), \textit{G-channel} comes into play. Notably, as particles serving \textit{G} remain static until around the \textit{140}th iteration, the diversity influenced by \textit{G-channel} experiences a sharp increase, surpassing that influenced by the \textit{non-G-channel}, thereby mildly warming up the algorithm's diversity. In the later stages of iteration, owing to the generation of a substantial number of $\textbf{\textit{P}}_n$s, the diversity influenced by \textit{non-G-channel} experiences a slight resurgence, perpetuating its exploration ability (particularly pronounced in multimodal problems, where the diversity exhibits more substantial growth). Meanwhile, the diversity influenced by the \textit{G-channel} diminishes due to the population aggregation guided by \textit{\textbf{G}}, showing exploitation ability.
	Further more, the diversity  influenced by  the both channels is greater on multimodal problems compared to  unimodal problems, which guarantees the generalization performance of the algorithm on different problems.  Furthermore, this assertion is also supported by the experimental findings detailed in section \ref{experiment}.
	
	\begin{algorithm}[!t]
		\renewcommand{\algorithmicrequire}{\textbf{Input:}}
		\renewcommand{\algorithmicensure}{\textbf{Output:}}
		\caption{Pseudocode of DCPSO-ABS}
		\label{VRD-PSO}
		\begin{algorithmic}[1]
			\Require Simple objective optimization problem with $f$; Search space $S^D$, $D$ is the number of decision variables; The population size $N$; the maximal number of function evaluations $FEs_{max}$;  refreshing gap $M$; 
			\Ensure  $\boldsymbol{G}$; 	\\
			/* Cell Division Initialization */
			\State 	$k_{max}=\lfloor FEs_{max}/N \rfloor$,  $k=0$,	$FEs=0$; 
			
			\For{1 $\leq$ n $\leq$ N}
			\State Randomly initialize $\boldsymbol{X}_n$ and $\boldsymbol{V}_n$, and evaluate $f(\boldsymbol{X}_n)$;
			\State Set $\boldsymbol{P}_n=\boldsymbol{X}_n$, and update $\boldsymbol{G}$;
			\EndFor
			\State $k=1$, $FEs=N$;
			\State Set that $\underset{non-G}{\boldsymbol{X}_n}$ and $\underset{G}{\boldsymbol{X}_n} $ are equal to $\boldsymbol{X}_n$, $ \underset{non-G}{\boldsymbol{V}_n}$ and $\underset{G}{\boldsymbol{X}_n}$ are qual to $\boldsymbol{V}_n$, $\alpha_n = 0$, $\beta_n=0$;
			
			/* Main Loop */
			\While{$k \leq k_{max} \&\& FEs \leq  FEs_{max}$}
			\State $k=k+1$;
			\For{1 $\leq$ n $\leq$ N}
			\State Update $\alpha_n$, $\beta_n$, $\boldsymbol{P}_n$, $\boldsymbol{G}$ and $FEs$ by Algorithm \ref{strategy 2};
			\EndFor
			\EndWhile
			\State \Return  $\boldsymbol{G}$.
		\end{algorithmic}
	\end{algorithm}

\textit{2) Computational Complexity:} For the proposed DCPSO-ABS algorithm, its computational cost mainly derives from initialization ($T_{ini}$) and loop ($T_{loop}$). it is easy to compute $T_{ini}$ as $O(D)$. 	For each sub-swarm in each $FEs$, the worst-case time complexity of $T_{loop}$, we can compute, is $O(D)$. Hence, the time complexity of the DCPSO-ABS algorithm is $O(D*FEs_{max})$, which is linear in $FEs_{max} $ and equal to the PSO evaluated in \cite{7271066}.	
	
	\section{Experiments Evaluation and  Discussion}\label{experiment}
	To illustrate the generalization performance of DCPSO-ABS algorithm,  Three kinds of experiments are designed. In the first type of experiments, we compare DCPSO-ABS algorithm with seven popular algorithms to illustrate  the  convergence performance of our algorithm on different benchmark functions. The second type of experiments confirms the stability and scalability of DCPSO-ABS by designing different numbers of $FEs_{max}$ and $D$. And the third type of experiments is the ablation experiment to be designed to investigate the the effect of two fundamental components, namely D-C and PDG, on  the performance of DCPSO-ABS.

	\subsection{ Experimental Setup} \label{bench mark}
	\textit{1) Benchmark Functions:} The CEC2013 \cite{liang2013problem} and CEC 2017 \cite{wu2017problem} test suites are  widely  used in evaluations. Specially, the 30 test benchmark functions in CEC 2017  with boundary constraints are all complex  to be often used to comprehensively assess algorithmic performance. However,  an official statement acknowledges that the Shifted and Rotated Sum of Different Power Function in CEC 2017 exhibits instability in higher dimensions, posing challenges in accurately evaluating algorithmic efficacy. Therefore, the remaining 57 functions are involved in our experiments, which are divided into three groups: Unimodal Functions, Simple Multimodal Functions, and Complex Multimodal Functions, as shown in Table S-I in the attached Supplementary Material.
	
	\textit{2) Parameter Settings of PSOs:} Seven state-of-the-art peer algorthms are  simultaneously compared with our proposed algorthm: HPSO-TVAC \cite{ratnaweera2004self}, CLPSO \cite{liang2006comprehensive}, OLPSO \cite{zhan2009orthogonal}, HCLPSO \cite{lynn2015heterogeneous}, EPSO \cite{lynn2017ensemble}, AWPSO \cite{liu2019novel}, and MAPSO \cite{wei2020multiple}. The parameter configurations of  these algorithms and the DCPSO-ABS algorithm are presented Table S-II in the attached Supplementary Material.
	
In addition, a statistical hypothesis test is applied to assess the significance of the differences in results. In this paper, a widely used nonparametric statistical hypothesis test, the  Wilcoxon signed-rank test, is employed at a significance level of $\alpha = 0.05$. In the evaluation, the symbols (+) and (-) indicate the DCPSO-ABS performs significantly better or worse than the compared algorithm, respectively, while the symbol (=) denotes that there is no significant difference  between DCPSO-ABS and the compared algorithm.
	
	\begin{table*}[!t]
		\centering
		\caption{comparison results  on unimodal functions (F$_1$-F$_7$) }
		\includegraphics[width=5.3in]{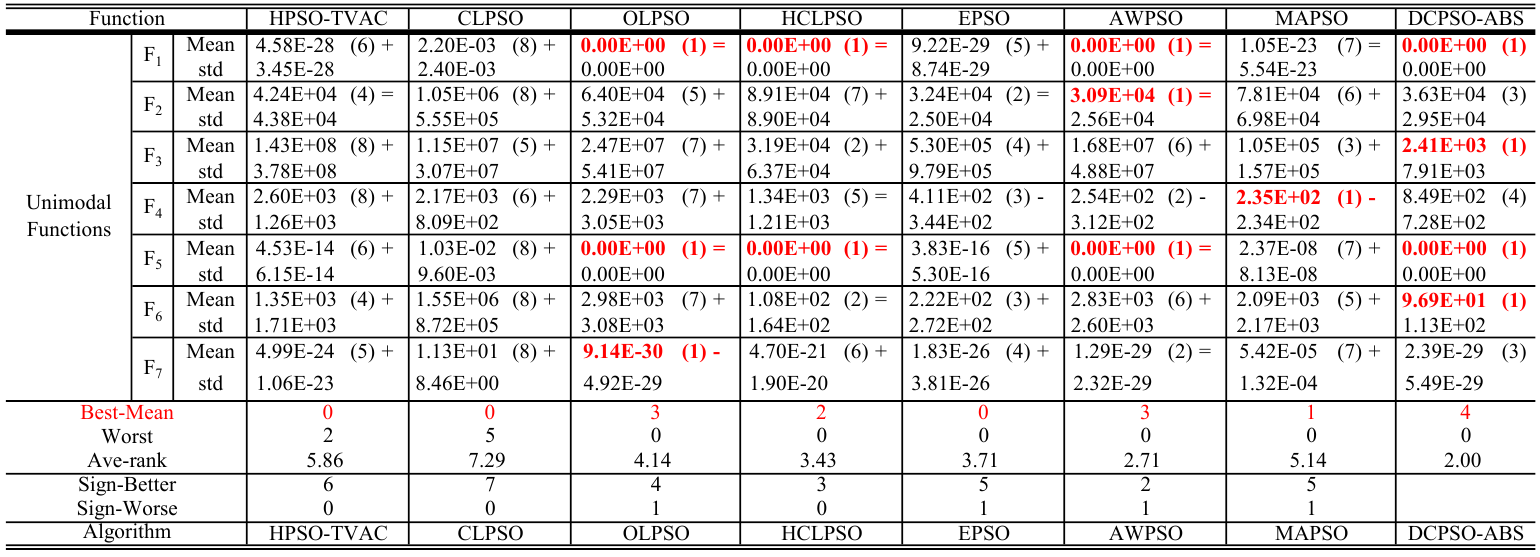}
		
		\label{comparison results of solution accuracy on unimodal functions (F$_1$-F$_7$)}
	\end{table*}

				\begin{table*}[!t]
		\centering
		\caption{comparison results on simple multimodal functions (F$_{8}$-F$_{29}$) }
		\includegraphics[width=5.3in]{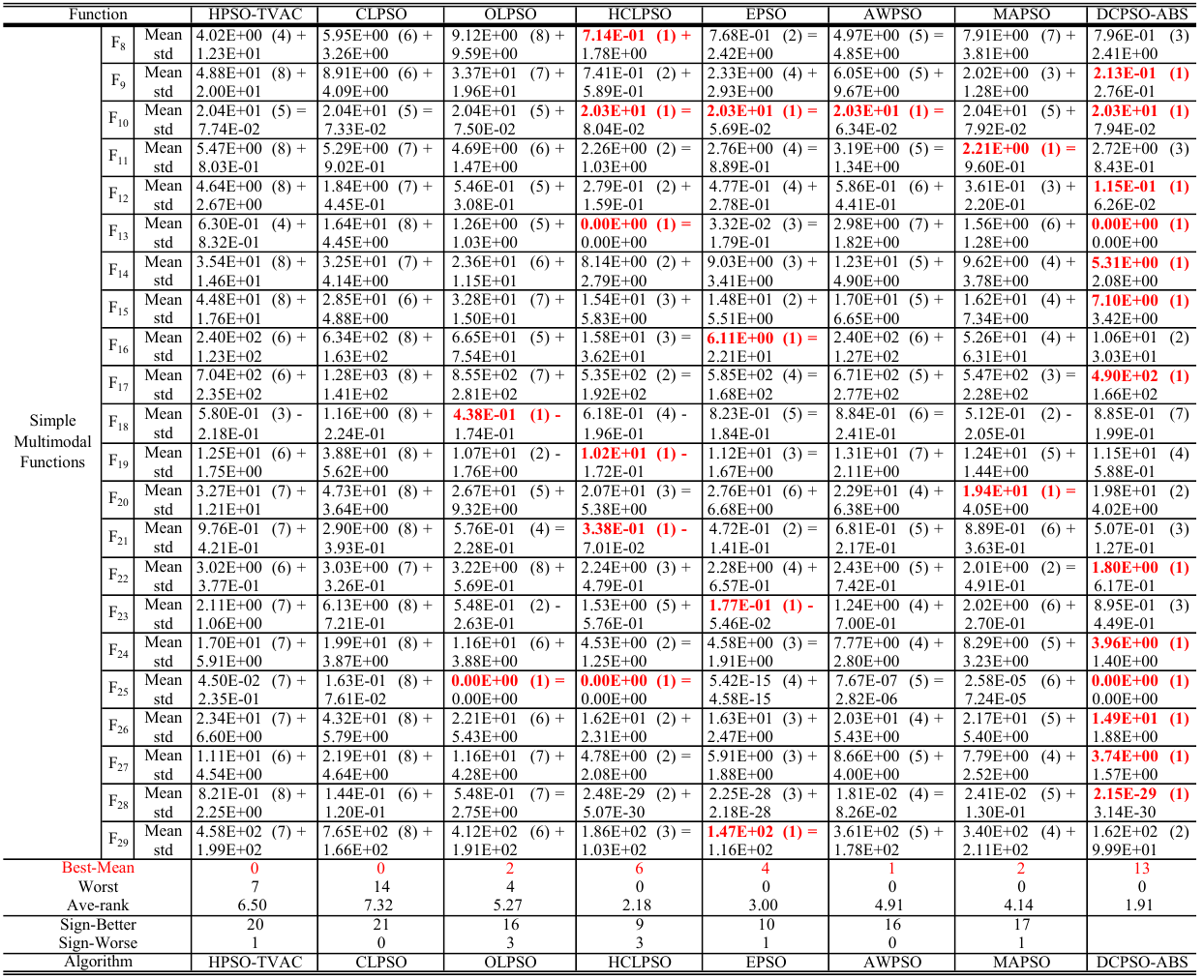}
		
		\label{comparison results of solution accuracy on simple multimodal functions (F$_{8}$-F$_{29}$)}
	\end{table*}
		
	\begin{table*}[t]
		\centering
		\caption{comparison results  on complex multimodal  functions (F$_{30}$-F$_{57}$) }
		\includegraphics[width=5.3in]{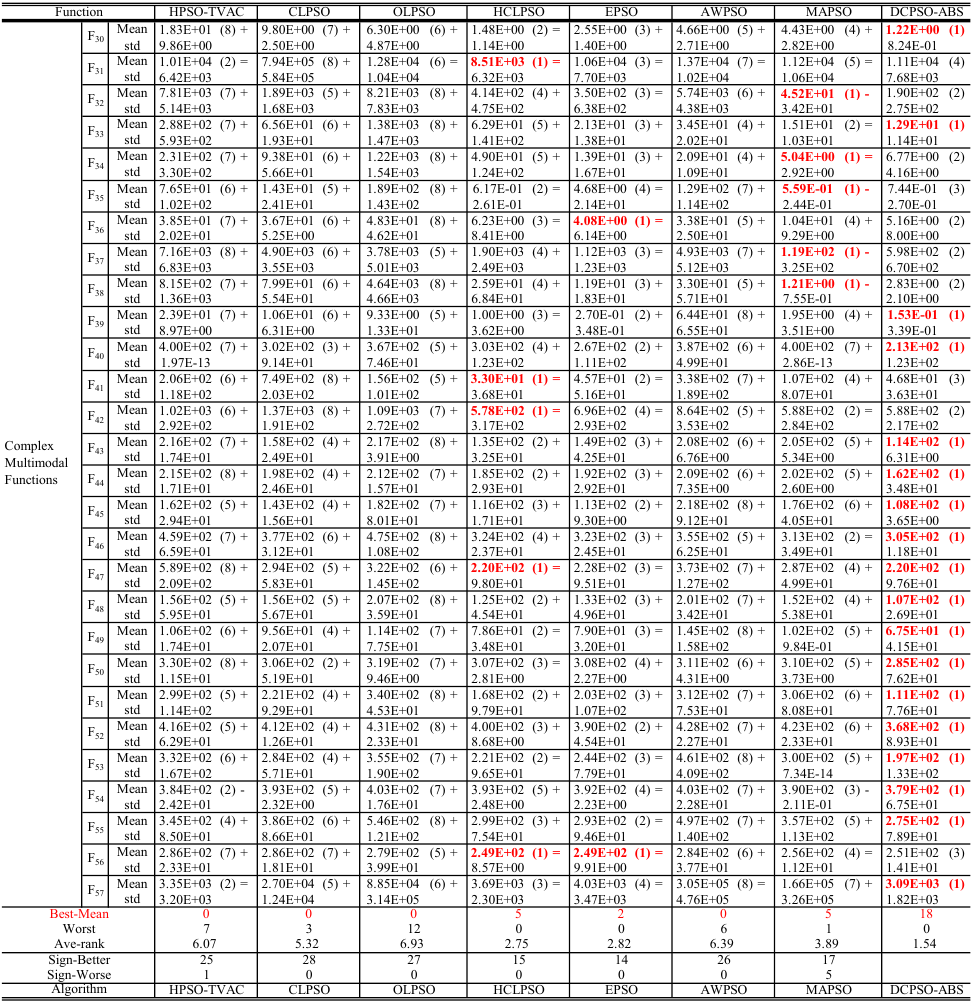}
		
		\label{comparison results of solution accuracy on hybrid  functions (F$_{30}$-F$_{57}$) }
	\end{table*}

	\subsection{Experimental Results of Convergence Performance} \label{Experimental Results of Performance and Generalizability}
	
\begin{figure*}	[t]
	\subfigure[]{
		\includegraphics[width=1.85in]{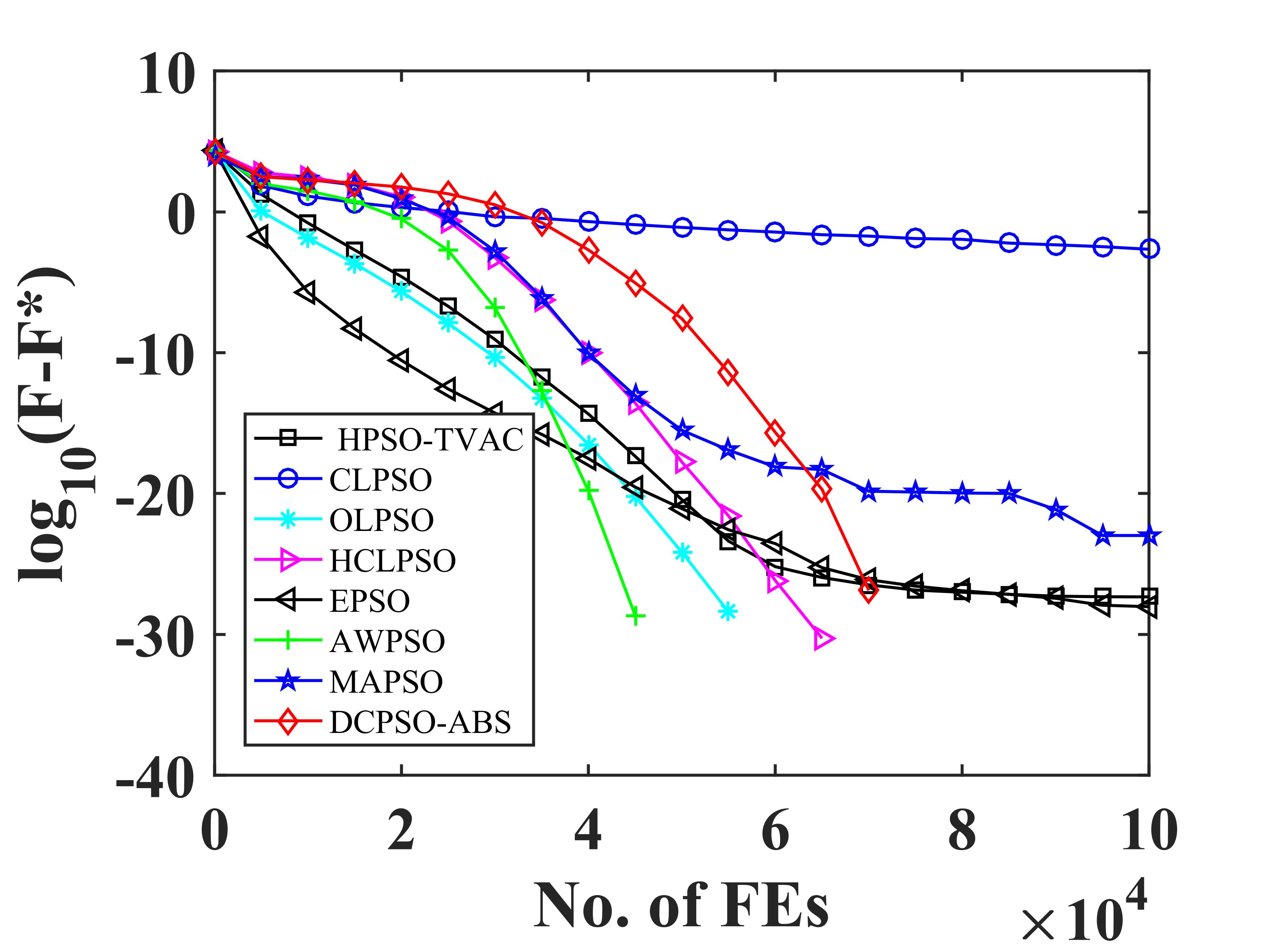}\\
	}%
	\hspace{-5.5mm}
	\subfigure[]{
		\includegraphics[width=1.85in]{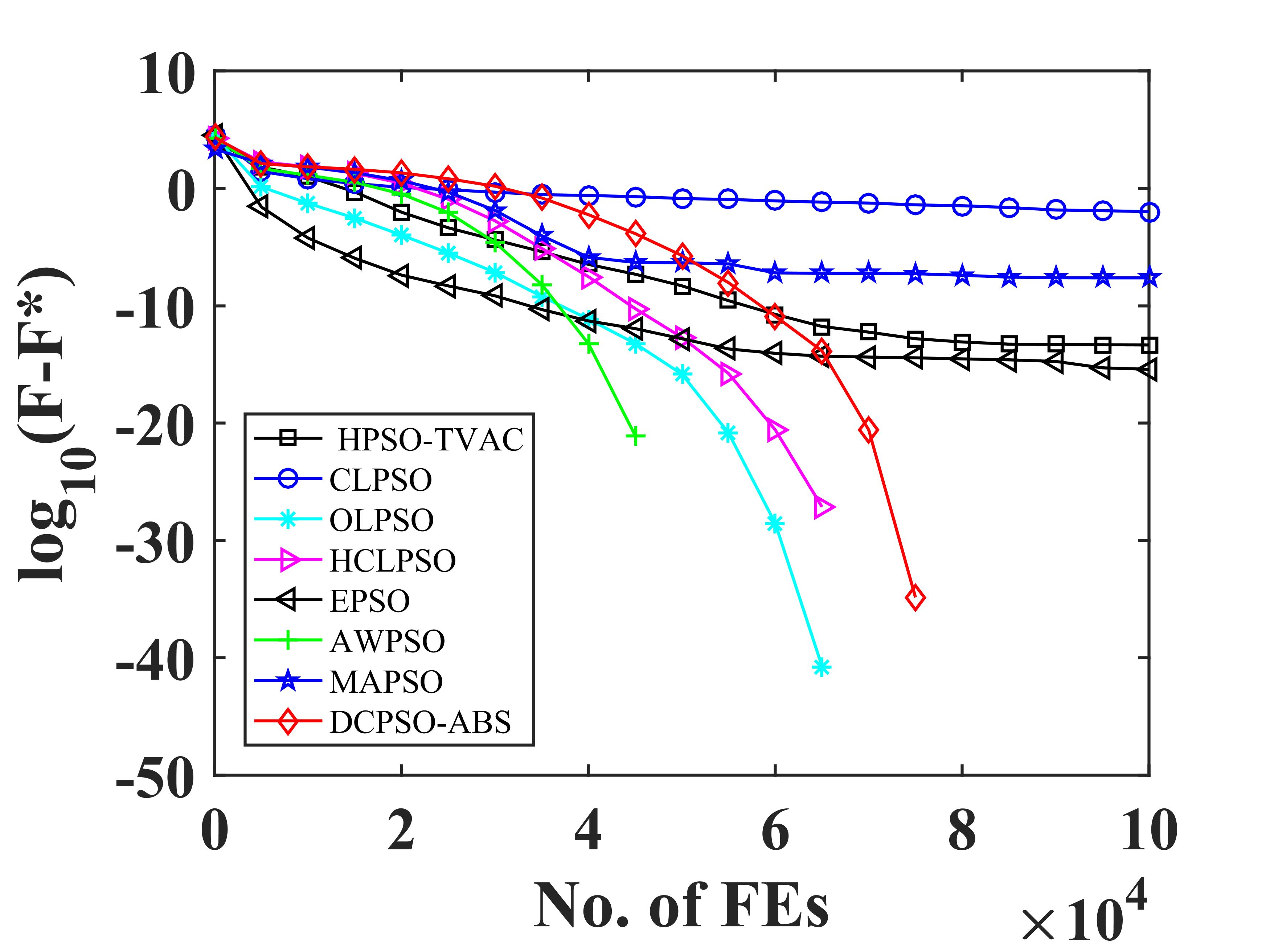}\\
	}	
	\hspace{-5.5mm}
	\subfigure[]{
		\includegraphics[width=1.85in]{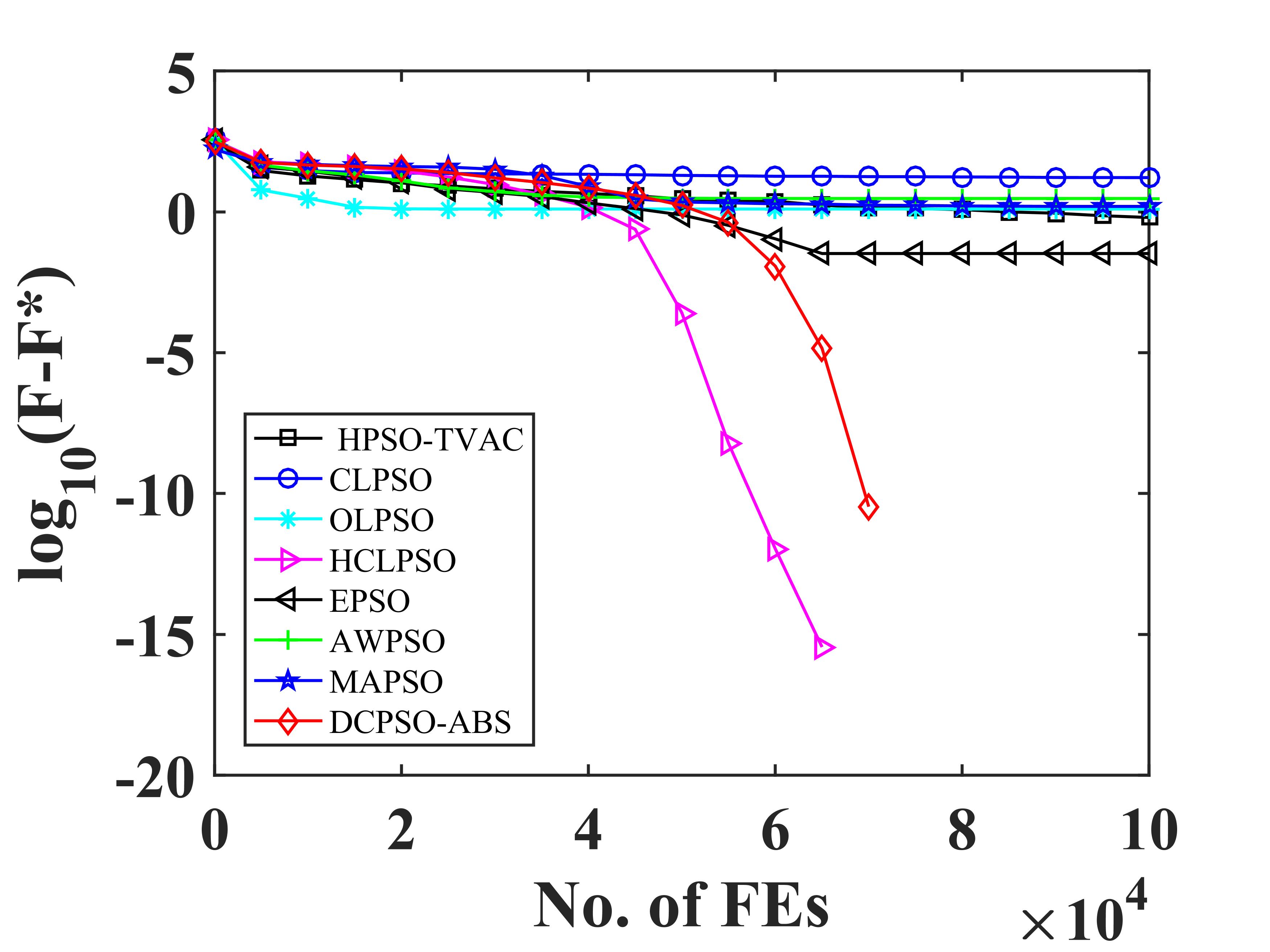}\\
	}%
	\hspace{-5.5mm}
	\subfigure[]{
		\includegraphics[width=1.85in]{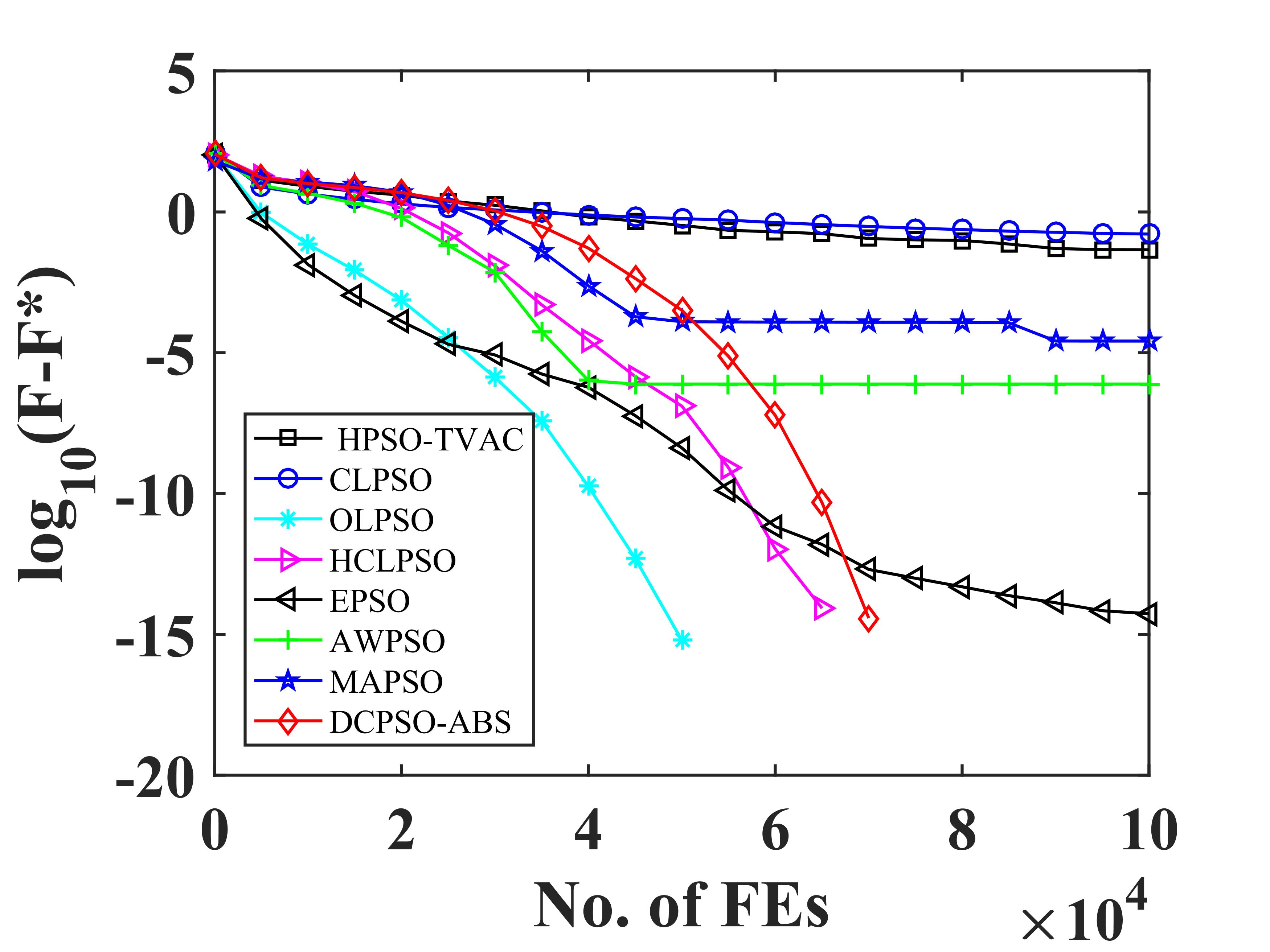}\\
	}%
	\hspace{-5.5mm}   
	\subfigure[]{
		\includegraphics[width=1.85in]{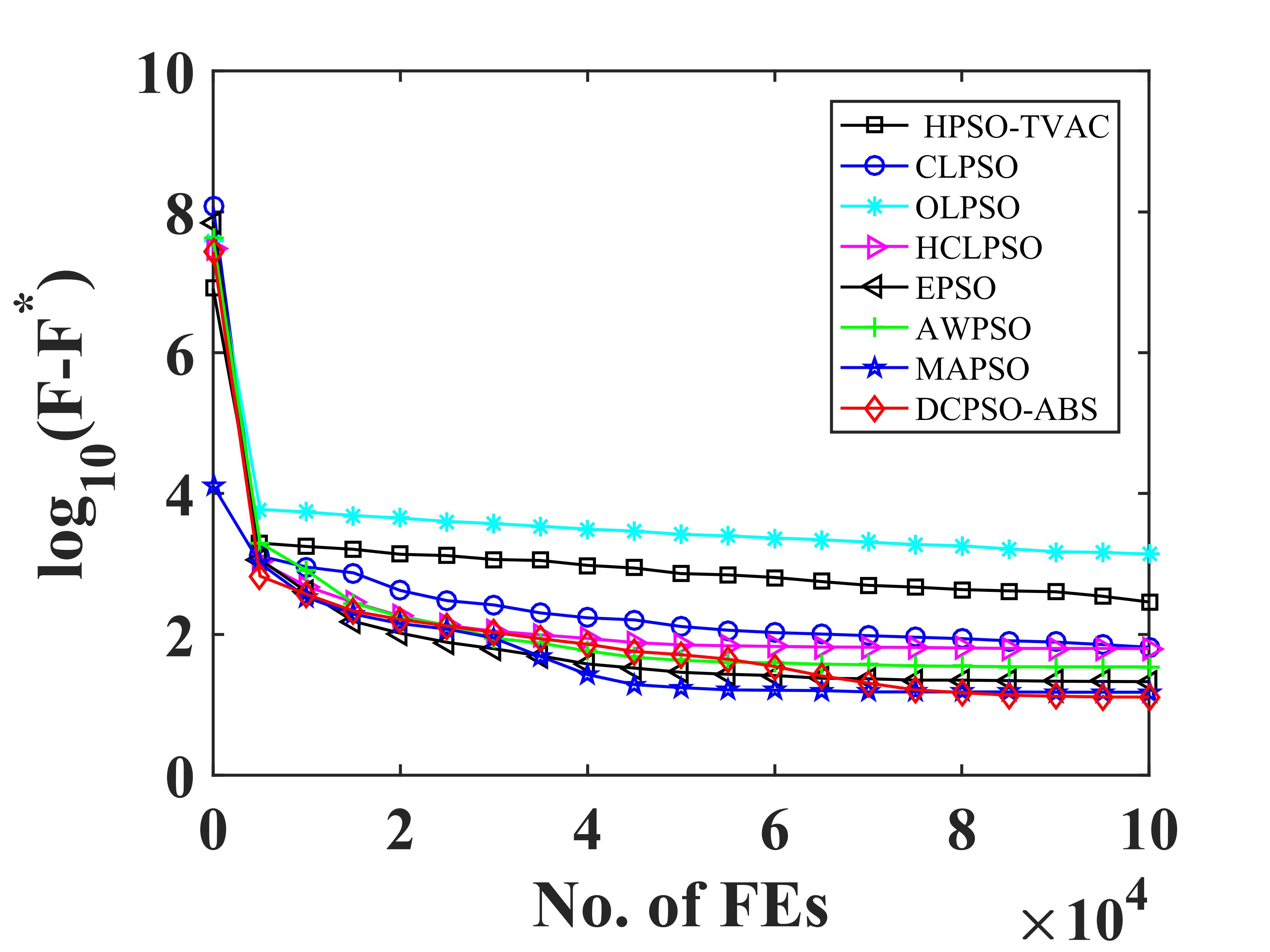}\\
	}%
	\hspace{-5.5mm}
	\subfigure[]{
		\includegraphics[width=1.85in]{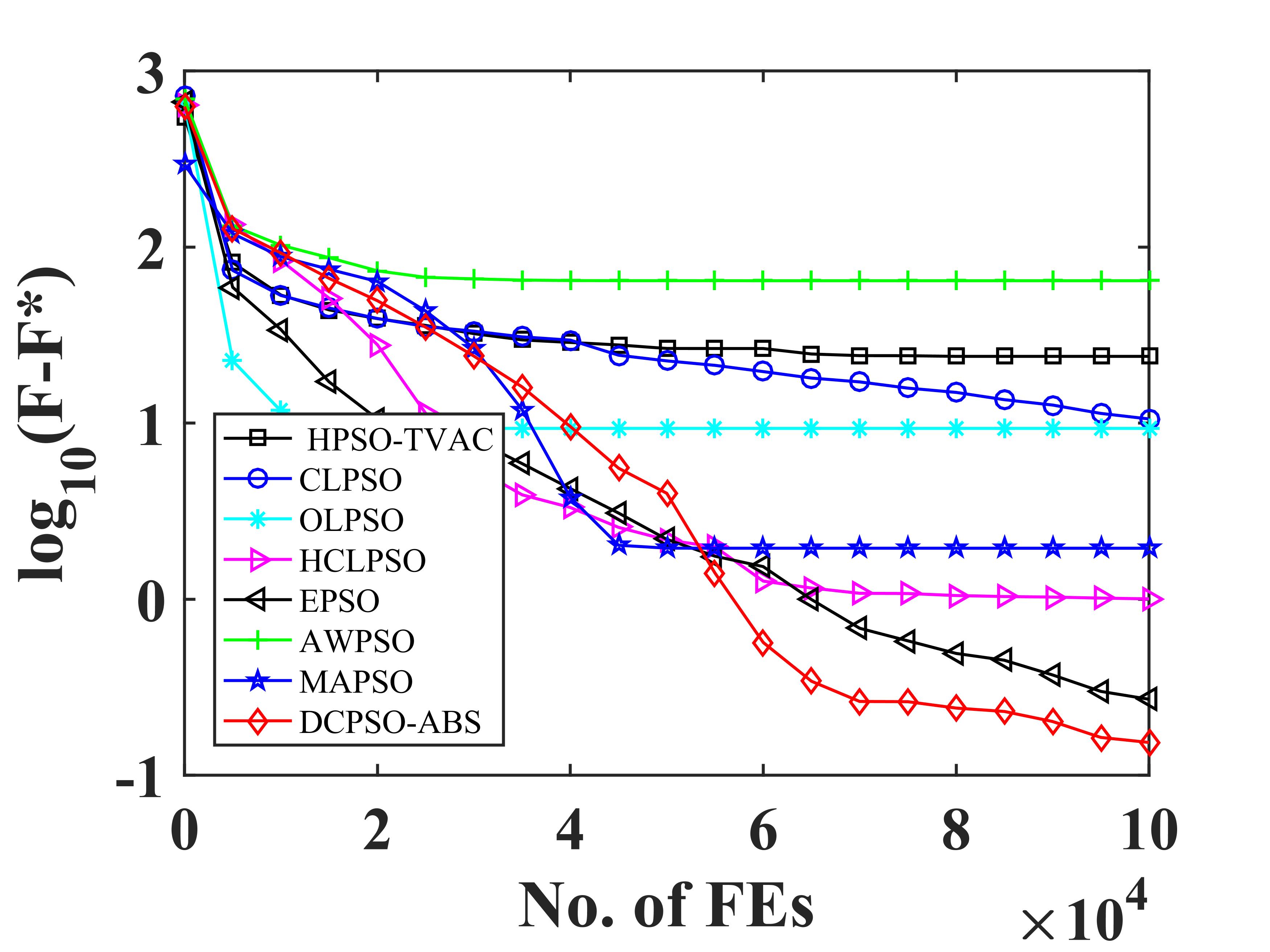}\\
	}	
	\hspace{-5.5mm}
	\subfigure[]{
		\includegraphics[width=1.85in]{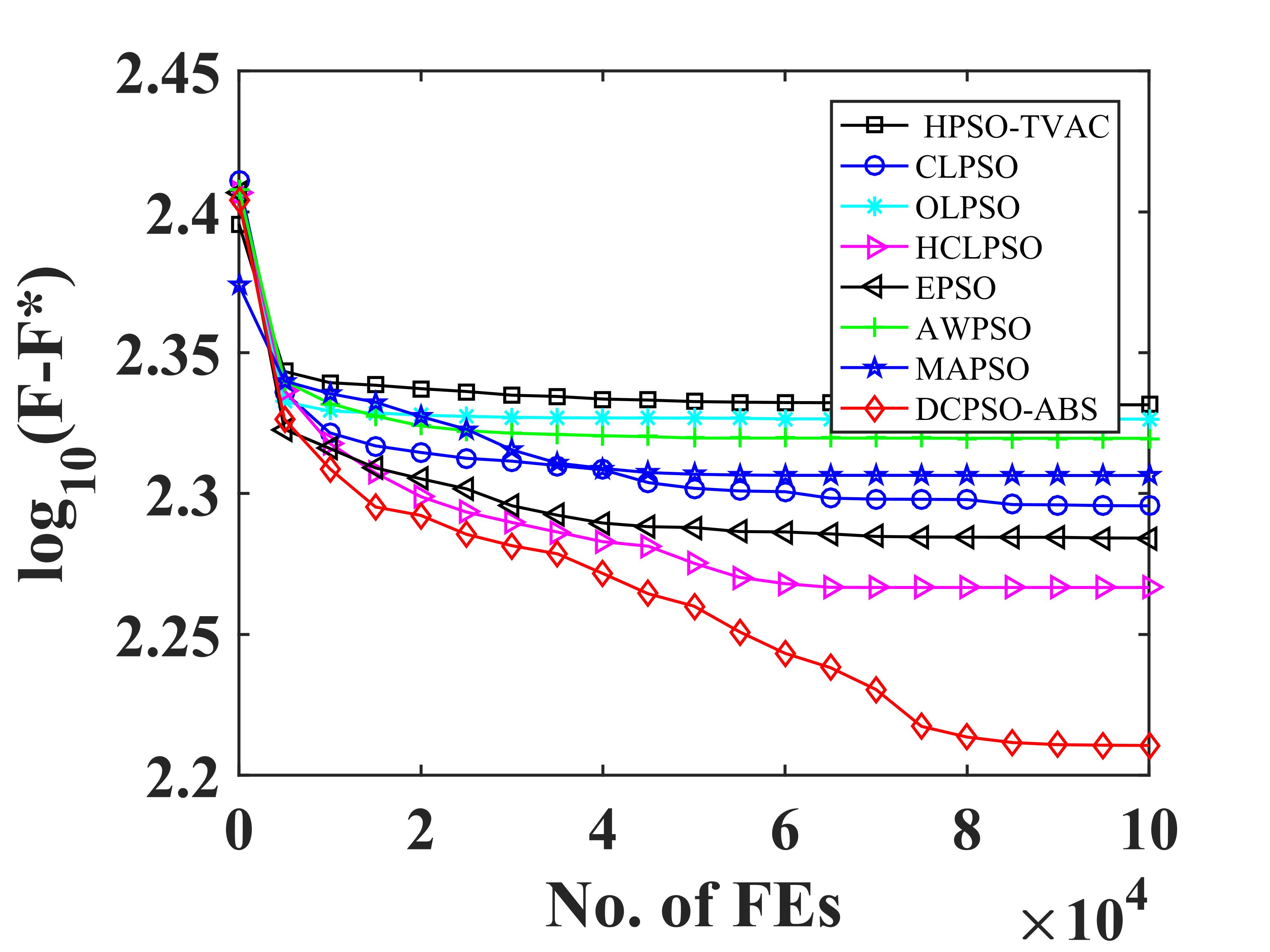}\\
	}%
	\hspace{-5.5mm}
	\subfigure[]{
		\includegraphics[width=1.85in]{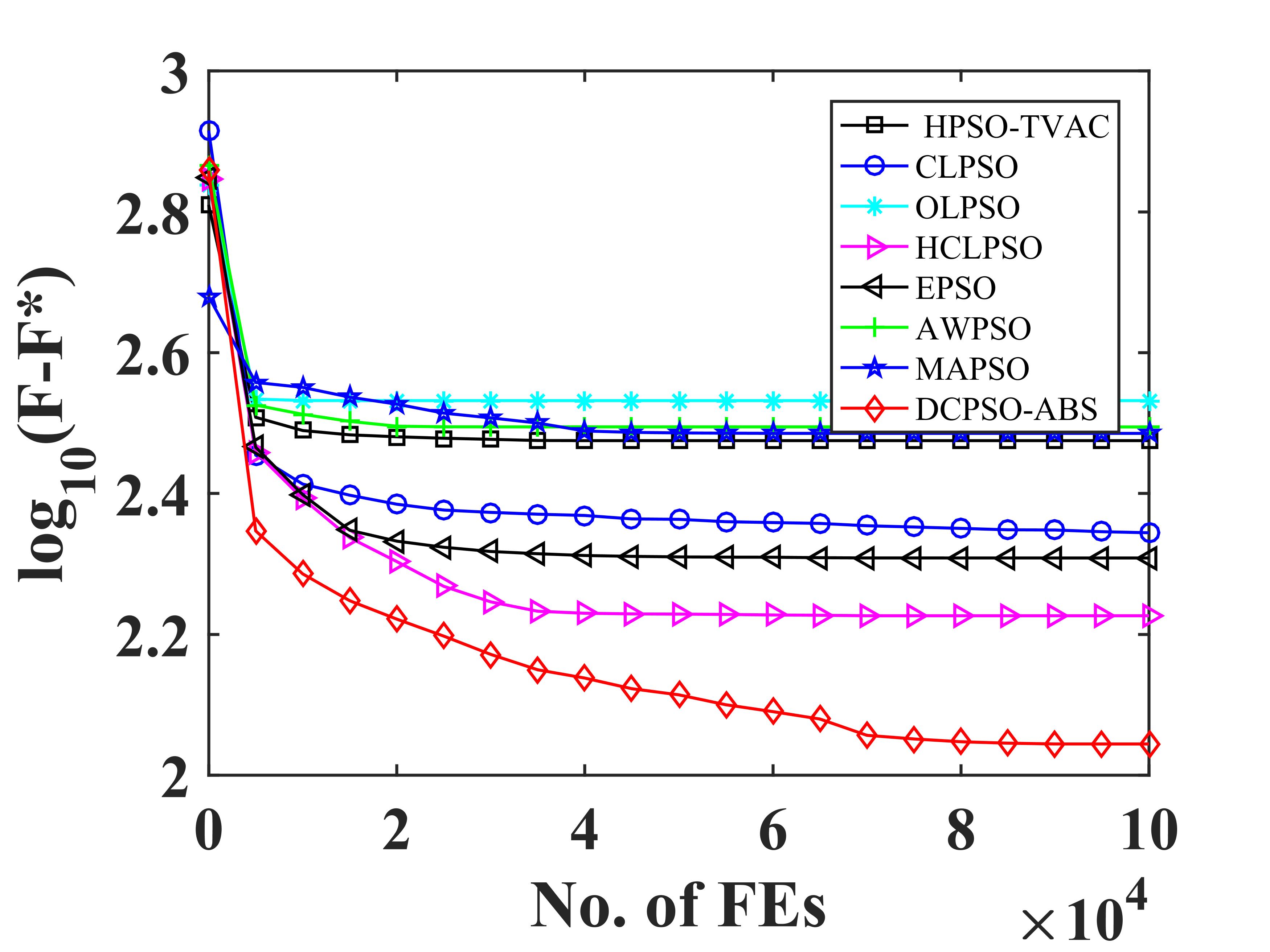}\\
	}%
	\centering
	\caption{Performance curve of 8 algorithms. (a) F$_1$. (b) F$_5$. (c) F$_{13}$. (d) F$_{25}$. (e) F$_{33}$. (f) F$_{39}$. (g) F$_{44}$. (h) F$_{51}$.}
	\label{Performance curve of 8 algorithms}
\end{figure*}

To ensure the fairness of the algorithm comparison, we maintain a fixed population size of 20 for all algorithms, which is a popular size for these compared algorithms\cite{lynn2015heterogeneous}, \cite{cao2018comprehensive}. Additionally, each algorithm undergoes test with the same number of function evaluations, set at $FEs_{max}=10,000\cdot D$, where \textit{D} is 10.  Each algorithm is executed independently 30 times on every benchmark function to acquire statistical results.  These results include the mean (Mean) of the error values, the standard deviations (std) of the error values, the ranks of the mean (rank), and the outcomes of the hypothesis test (Sign-Better / Sign-Worst), are displayed in Table \ref{comparison results of solution accuracy on unimodal functions (F$_1$-F$_7$)}, Table \ref{comparison results of solution accuracy on simple multimodal functions (F$_{8}$-F$_{29}$)}, and Table \ref{comparison results of solution accuracy on hybrid functions (F$_{30}$-F$_{57}$) }, where the best Mean results are marked in red bold. In these Tables,  the mean of ranks (Ave-rank) is calculated, and the Best-Mean and Worst are pronounced to represent the number of the best or worst Mean. Moreover, we illustrate the performance curves of the eight algorithms in the eight representative functions,  shown in  Fig. \ref{Performance curve of 8 algorithms}.

Table \ref{comparison results of solution accuracy on unimodal functions (F$_1$-F$_7$)} presents the results on Unimodal Functions (F$_1$-F$_7$) of the eight algorithms. In this Table, it is clear that CLPSO has dramatical challenges to achieve superior solution accuracy. It is possibly because the CL strategy in CLPSO is to enhance the diversity while weakening the influence of \textit{\textbf{G}} unintentionally.
  Although OLPSO and AWPSO both exhibit the capability to achieve high solution accuracy through their innovative strategies, they are yet slightly inferior compared to DCPSO-ABS, no matter on Best-Mean or Ave-rank. Moreover, the hypothesis test results also  indicate that our proposed algorithm demonstrates significant effectiveness.  These evidences explain that our proposed algorithm has better convergence performance on Unimodal Functions. Additionally, it is worth mentioning that DCPSO-ABS also secures the top rank on both F$_3$ and F$_6$, which are two derivative functions of Bent Cigar Function. This outcome suggests that DCPSO-ABS exhibits strong generalizability when facing various operations of a single unimodal function, such as rotated and shifted.

The experimental results on Simple Multimodal Functions (F$_8$-F$_{29}$) are illustrated in Table \ref{comparison results of solution accuracy on simple multimodal functions (F$_{8}$-F$_{29}$)}. According to the Best-Mean,  DCPSO-ABS secures the most number of times about first rank, more than twice as much as HCLPSO at second place.  Although OLPSO and AWPSO perform well on Unimodal Functions, underperform on Simple Multimodal Functions, possibly resulting from the unreasonable controllment of the \textit{\textbf{G}}'s behavior.
Moreover, our proposed algorithm demonstrates substantial superiority over others in terms of Ave-rank and the hypothesis test results. All results in Table III imply that our proposed algorithm has remarkable convergence performance.  Further considering two sets, $\{$F$_{13}$, F$_{14}$, F$_{24}$$\}$ (variants of the Rastrigin Function) and $\{$F$_{16}$, F$_{17}$, F$_{29}$$\}$ (variants of the Schwefel Function), our proposed algorithm demonstrates superior results compared to other algorithms. This suggests that DCPSO-ABS may possess a robust generalization capability across various operations (rotated, rotated, and shifted) for  simple multimodal functions.

\begin{figure*} [!]
	\subfigure[]{
		
		\includegraphics[width=1.8in]{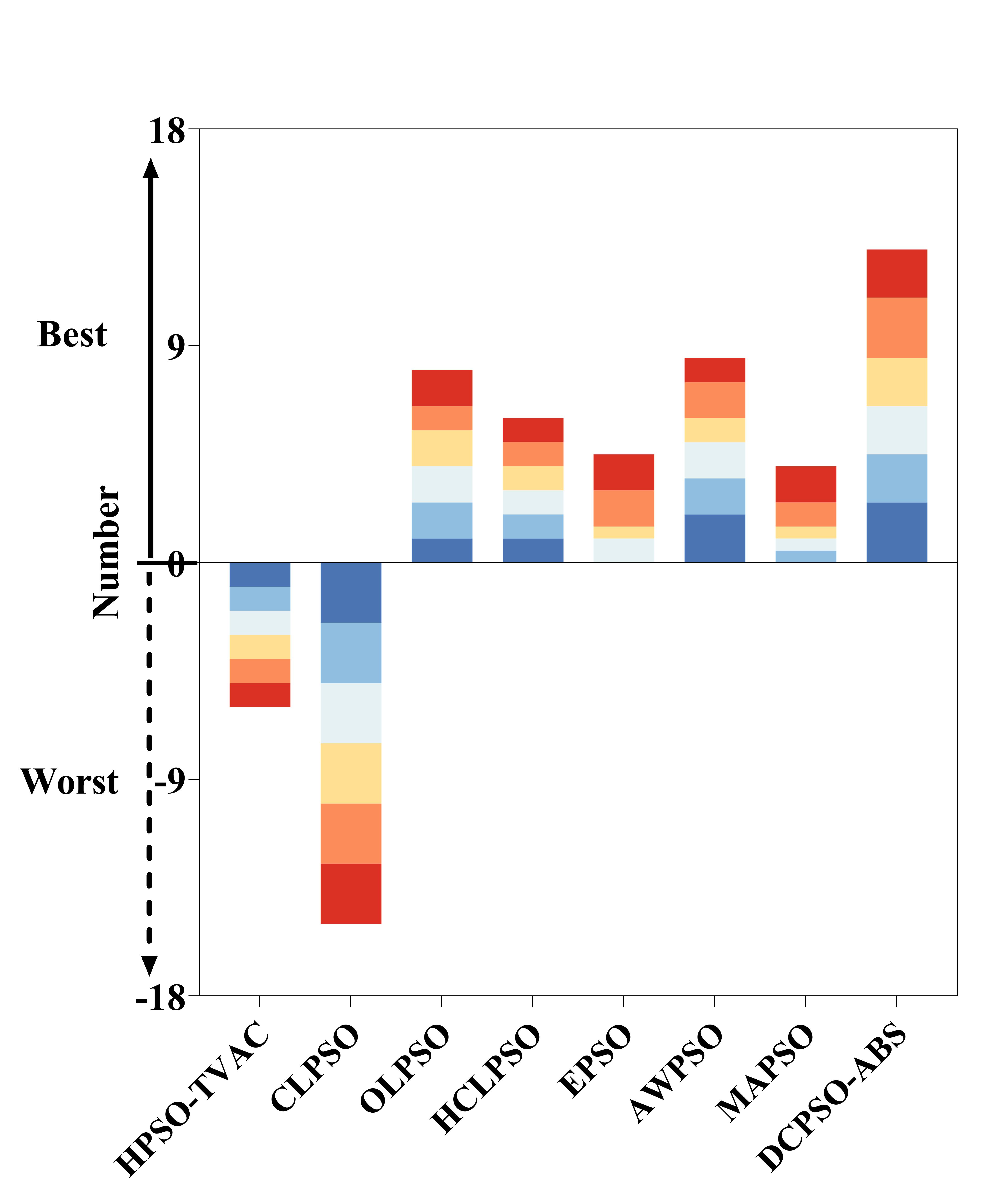}\\
	}%
	\hspace{6mm}
	\subfigure[]{
		\includegraphics[width=1.7in]{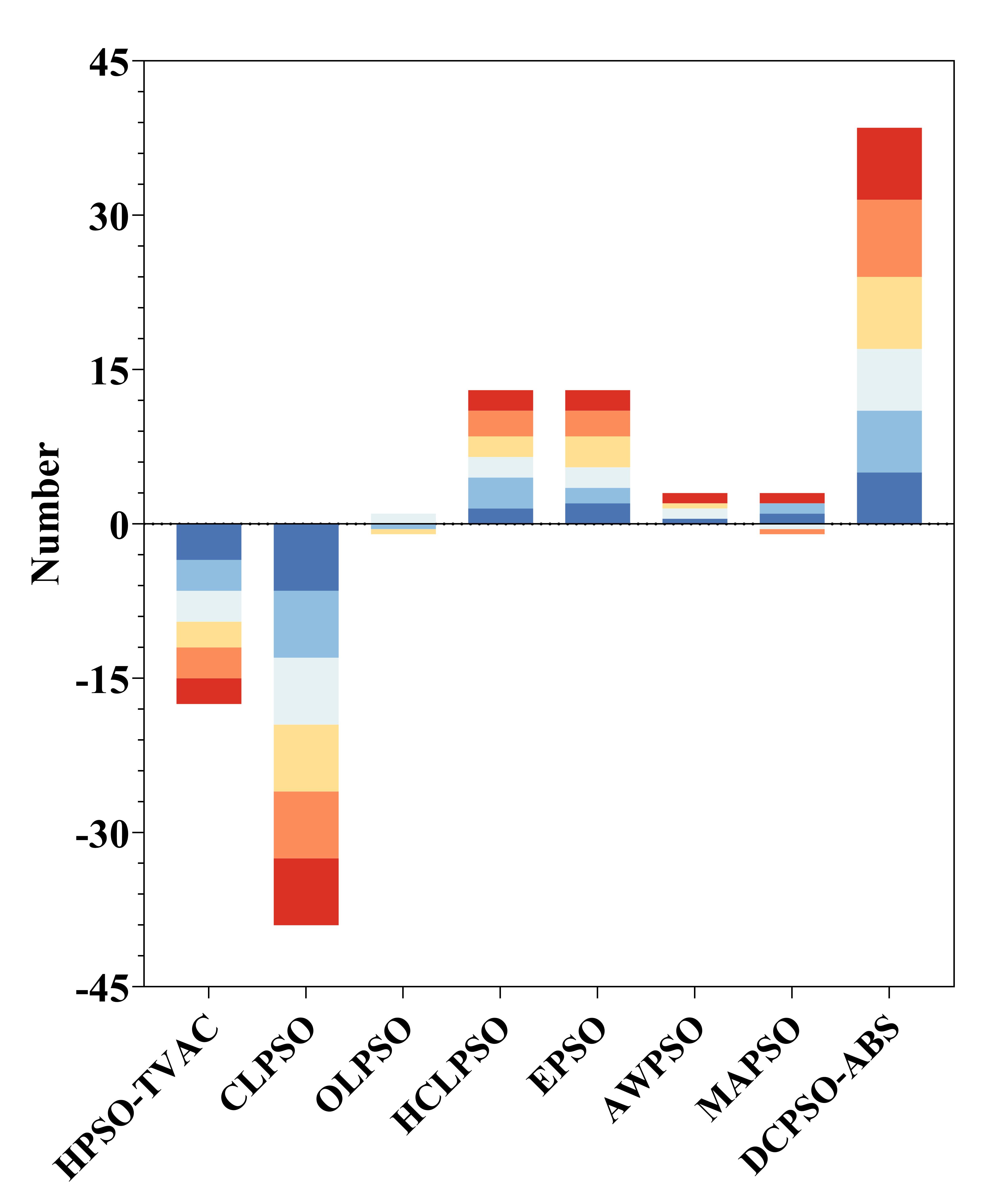}\\
	}
	\hspace{6mm}
	\subfigure[]{
		\includegraphics[width=2.05in]{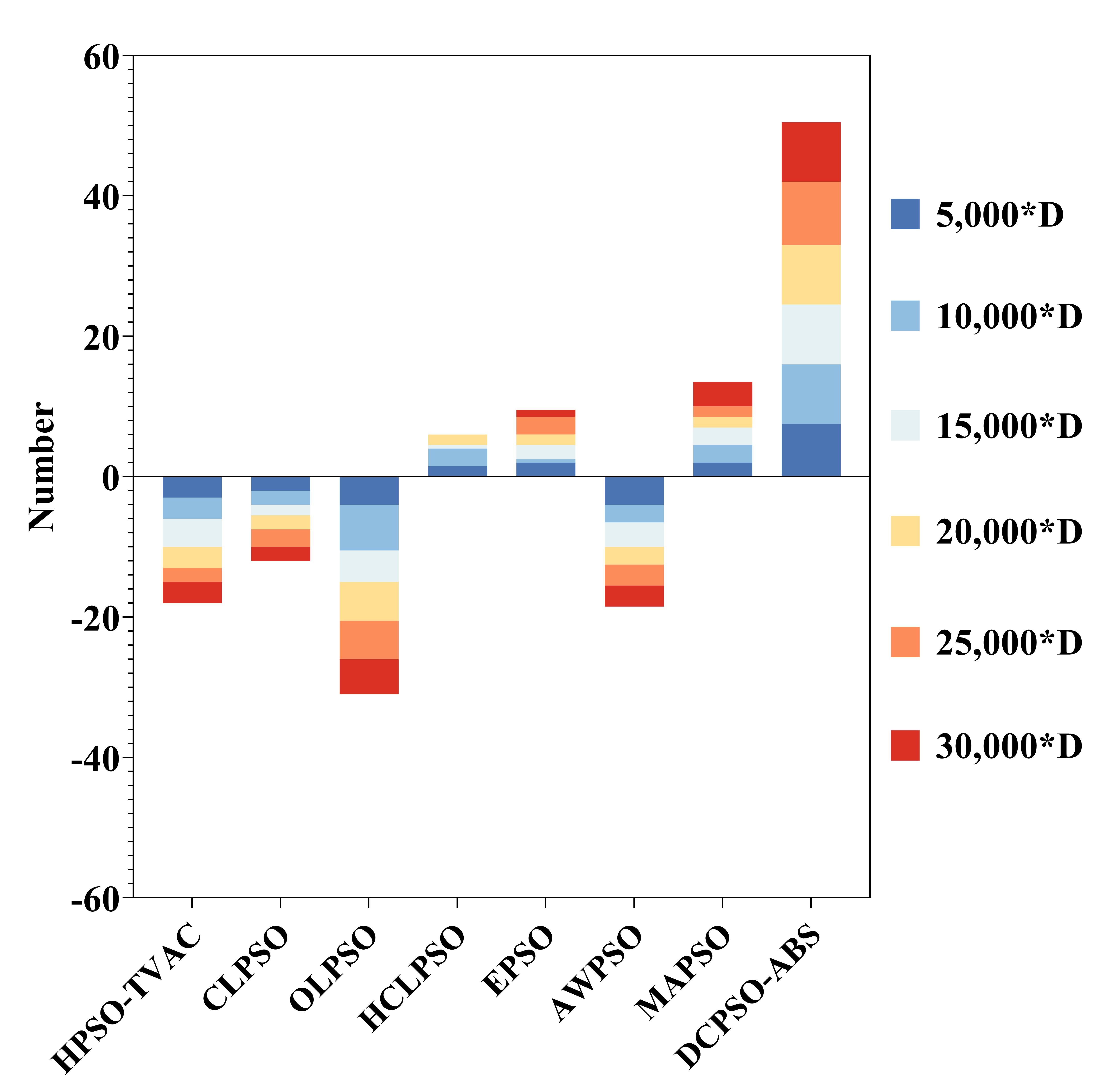}\\
	}%
	\centering
	\caption{
		Comparison results under different $FEs_{max}$. (a) Unimodal Functions.   (b) Simple Multimodal Functions. (c) Complex Multimodal Functions.}
	\label{Performance of 8 algorithms affected by $FEs_{max}$  on (a) Unimodal Functions,   (b) Simple MultimodalFunctions, (c) Hybrid Functions, (d) Composition Functions.}
\end{figure*}

Table \ref{comparison results of solution accuracy on hybrid functions (F$_{30}$-F$_{57}$) } presents the experimental results on Complex Multimodal Functions (F$_{30}$-F$_{57}$), which are primarily used to simulate real-world optimization problems.  According to Worst in Table \ref{comparison results of solution accuracy on hybrid functions (F$_{30}$-F$_{57}$) }, it can be found that the convergence performance of CLPSO has warmed up on these functions than that on Unimodal Functions and Simple multimodal Functions, which is mainly due to the CL strategy.  Furthermore, HCLPSO based on the CLPSO, further considering the convergence feature of \textit{\textbf{G}},  shows better performance than CLPSO, even taking the second place in  Ave-rank.  However, due to a lack of the comprehensive controllment of \textit{\textbf{G}}'s behavior, the performance of HCLPSO lags far behind the DCPSO-ABS we proposed.  Moreover, the hypothesis testing results also imply the superiority of the DCPSO-ABS algorithm over the other seven algorithms within this set of functions.  It is also found that OLPSO and AWPSO still underperform on these functions, while MAPSO performs better on this Functions than that on Unimodal and Simple Multimodal Functions.  
Jointly considering the Table \ref{comparison results of solution accuracy on unimodal functions (F$_1$-F$_7$)}, Table \ref{comparison results of solution accuracy on simple multimodal functions (F$_{8}$-F$_{29}$)} and Table \ref{comparison results of solution accuracy on hybrid  functions (F$_{30}$-F$_{57}$) }, we find that as the difficulty of the function increases, the  convergence performance of OLPSO and AWPSO gets worse, while MAPSO gets better. It indicates that OLPSO and AWPSO are biased towards \textit{Ei} regarding to the balance of \textit{Er} and \textit{Ei}, while MAPSO is biased towards \textit{Er}. It's worth mentioning that DCPSO-ABS consistently remarkably performs  in these three categories of Functions, implying that our proposed algorithm provides a better balance of \textit{Er} and \textit{Ei} for various problems.

\begin{table*}[t]
	\centering
	\caption{ comparison results of eight algorithms under different $D$ }
	\includegraphics[width=5.3in]{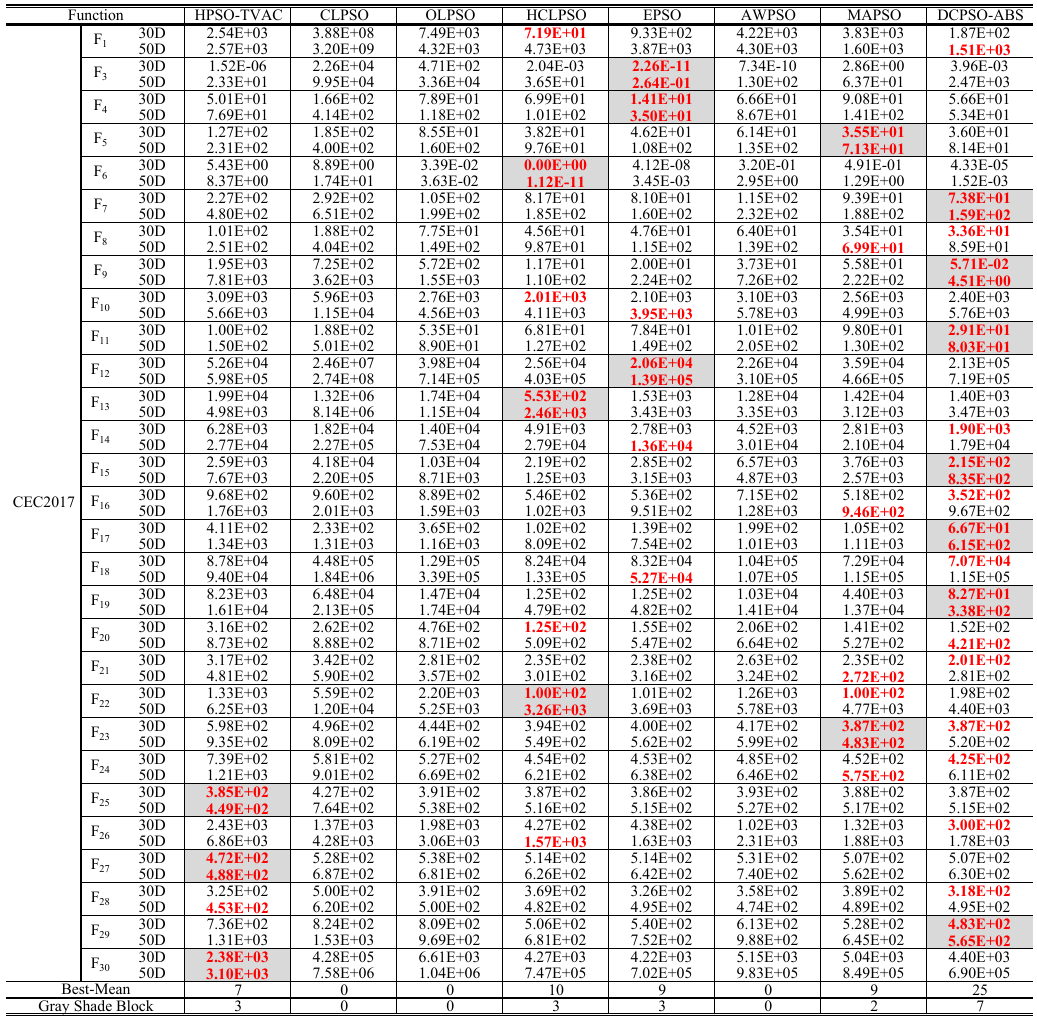}
	
	\label{inluence of D}
\end{table*}

   Furthermore, we illustrate the performance curves of eight algorithms in some representative functions, shown in Fig. \ref{Performance curve of 8 algorithms}. It is obviously observed that AWPSO shows the fastest convergence speed, followed by OLPSO in Fig. \ref{Performance curve of 8 algorithms} (a)-(b), while they  perform worse in Fig. \ref{Performance curve of 8 algorithms} (e)-(h), further confirming that they are biased towards \textit{Ei} regarding the balance of \textit{Er} and \textit{Ei}.  Moreover, what is more abrupt is that OLPSO finds the optimal solution while AWPSO does not in Fig. \ref{Performance curve of 8 algorithms} (d).  It may be derived from the fact that orthogonal learning strategy in OLPSO helps find a better direction by constructing a much promising and efficient exemplar \cite{zhan2009orthogonal},  while increasing the diversity to a certain extent.  This fact is also possibly the reason that the convergence speed of OLPSO is slightly slower than AWPSO. This obeys the free lunch theorem \cite{wolpert1997no}. Moreover, we find that DCPSO-ABS although shows a slightly slower convergence speed at the functions where the optimal solutions are found, gains the same optimal solution as the compared algorithms,  displayed in  Fig. \ref{Performance curve of 8 algorithms} (a)-(d). Additionally, Fig. \ref{Performance curve of 8 algorithms} (e)-(h) illustrate that our proposed algorithm exhibits better convergence effectiveness at the functions where  the optimal solutions are not found by these eight algorithms. These phenomena occur because we esteem the \textit{non-G-channel} and weaken the influence of \textbf{\textit{G}} in pre-iteration period to gain larger diversity, and maintain the convergence feature of \textit{\textbf{G}} in the later iteration period to pursue the accuracy of solutions. Intuitively, we get significantly improved convergence performance on most problems at only the cost of a slightly slower convergence speed, which  is acceptable in practical application and also complies with the Free Lunch Theorem.  In conclusion, when facing a variety of problems, our proposed algorithm shows strong generalization performance, which  mainly  thanks to the effective management of \textbf{\textit{P}} and \textit{\textbf{G}}'s behaviors by the DC framework and the ABS strategy.

	\subsection{Experimental Results of Stability and Scalability}
	
	It is a well-established fact that the performance of an algorithm is influenced by the number of $FEs_{max}$ and the dimensionality \textit{D}. Generally, performance enhances with larger \textit{FEsmax} and deteriorates with increased dimensions.  Hence, assessing the stability of algorithmic performance across varying \textit{FEsmax} and its scalability concerning diverse \textit{D} values are crucial analytical focal points.

	\textit{1) Stability on Different $FEs_{max}$:} In this section, six cases are designed with $FEs_{max} = 5,000\cdot D, 10,000\cdot D, 15,000\cdot D, 20,000\cdot D, 25,000\cdot D, 30,000\cdot D$, with $D=10$. And 30 independent experiments are conducted for each algorithm. In each case,  the occurrences of the best and worst outcomes of each algorithm are counted. The results  are illustrated in Fig. \ref{Performance of 8 algorithms affected by $FEs_{max}$  on (a) Unimodal Functions,   (b) Simple MultimodalFunctions, (c) Hybrid Functions, (d) Composition Functions.}. In the graphical representations, the upper bars represent the number of Best, while the lower bars denote the  number of Worst. 
	
Specifically, it is obvious that  the height of blocks of our proposed algorithm is highest both in terms of all cases of $FEs_{max}$ and each case of $FEs_{max}$, followed by AWPSO and OLPSO,	in  Fig. \ref{Performance of 8 algorithms affected by $FEs_{max}$  on (a) Unimodal Functions,   (b) Simple MultimodalFunctions, (c) Hybrid Functions, (d) Composition Functions.} (a), and that CLPSO appears in  Worst. It illustrates the strong performance of our proposed algorithm on Unimodal Functions. Moreover, the results on Simple Multimodal Functions display a rapid loss of the performance of OLPSO and AWPSO, and that OLPSO and MAPSO even appear at Worst, indicating their strong reliance on $FEs_{max}$. Additionally,  Fig. \ref{Performance of 8 algorithms affected by $FEs_{max}$  on (a) Unimodal Functions,   (b) Simple MultimodalFunctions, (c) Hybrid Functions, (d) Composition Functions.} (c) figures that the height of blocks of our proposed algorithm is extremely high, indicating that DCPSO-ABS has excellent performance on Complex Multimodal Functions. Moreover, the performance of OLPSO and AWPSO continues to deteriorate, while the performance of CLPSO has warmed up.  These results are consistent with the experimental results in Section \ref{Experimental Results of Performance and Generalizability}. Meanwhile, as the difficulty of the functions increases, the performance of these algorithms is trending downwards, while MAPSO trending upwards, probably due to his various adaptive strategies. Nevertheless, our algorithm  always shows strong performance on different Functions. It needs to be emphasised that the difference of DCPSO-ABS between different  $FEs_{max}$ is less than that of the seven compared algorithms on different Functions. It clarifies that DCPSO-ABS has strong stability  and further validates the importance of reasonably managing  \textit{\textbf{P}} and \textit{\textbf{G}}'s behaviors.

	\textit{2) Scalability on Different D:} The functions within the CEC 2017 test set  are complex and bounded, usually applied to assess the scalability of the algorithms. In this section, we design two cases: $D=30$ and $D=50$. In each case, each algorithm is operated independently 30 times under $FE_S=10,000 \cdot D$. The results are illustrated in  Table \ref{inluence of D}, where the gray shaded blocks in Table \ref{inluence of D} denote instances where an algorithm performs the best for a given function under both 30-D and 50-D.  As can be seen from Table \ref{inluence of D}, according to the results of Best-Mean, HPSO-TVAC contrarily holds better performance compared with the foregoing experiments, while CLPSO and OLPSO perform worse. It is mainly because  as the  dimension increases, the increasing number of particles improves the performance of HPSO-TVAC, while slowing the convergence speed of CLPSO and OLPSO, which is also the main reason why we modify the CL strategy in this paper. Besides, it can be found that DCPSO-ABS takes the top place, followed by  HCLPSO, EPSO and MAPSO, and that the number of DCPSO-ABS in Best-Mean is more twice than them. These results exhibit that our proposed algorithm still keeps the higher performance and competitiveness  on higher dimensional problems.  Furthermore, the number of shaded blocks associated with the DCPSO-ABS algorithm markedly outstrips the occurrences observed in the remaining seven algorithms, confirming the robust scalability inherent in our algorithm.

	\begin{table}[t]
	\centering
	\caption{ Comparison Results of DCPSO-ABS${_\text{p}}$,PSO-ABS${_\text{non-G-channel}}$, PSO-ABS${_\text{G-channel}}$, and DCPSO-ABS }
	\includegraphics[width=2.9in]{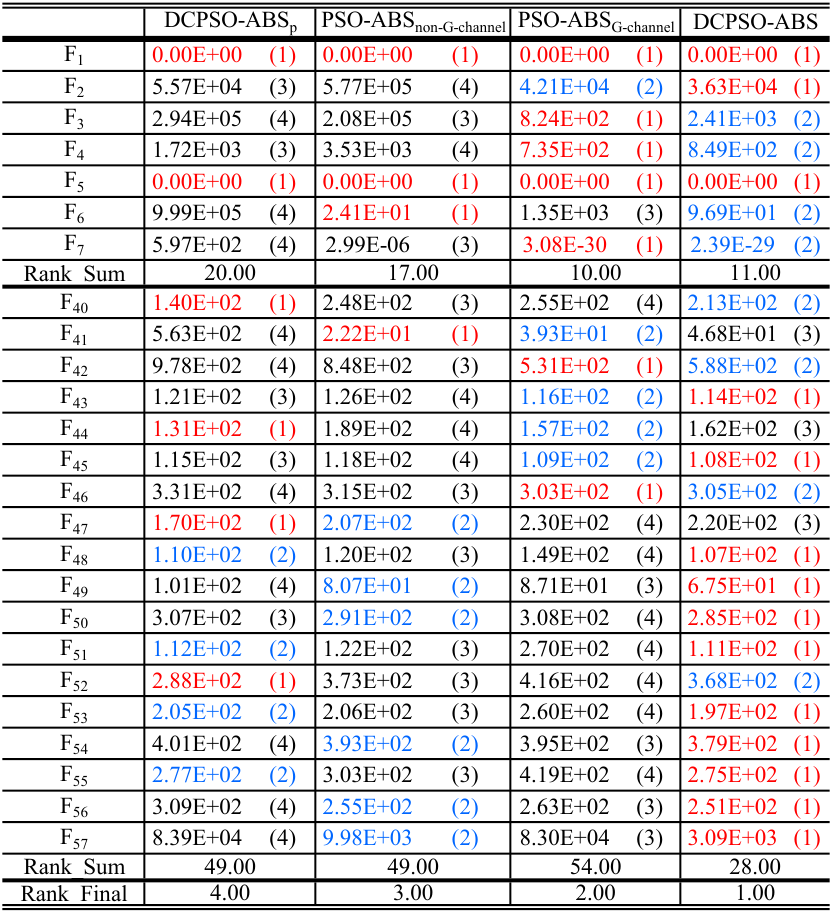} 	
	\label{strategy 1+2 influence performance}
\end{table}

	\subsection{Results of Ablation Experiment}
Through a series of foregoing research of DCPSO-ABS, we have investigated the effects of a series of controls for \textit{\textbf{P}} and \textit{\textbf{G}}'s behaviour on the balance of \textit{Er} and \textit{Ei} in section \ref{DCPSO-ABS}. In addition, D-C and PDG are worth discussing as foundations for the proposed algorithm. The former aids to regulate the two behaviors of \textbf{\textit{P}}  and \textbf{\textit{G}}, while the latter improves essential exploration ability. To evaluate the impact of the two components on the performance of DCPSO-ABS, a series of  experiments is conducted in this section. All functions under consideration are tested with $D = 10$, and all involved algorithms are computed with $FEsmax = 10,000 \cdot D$. In order to analyze the influence of PDG, we introduce the DCPSO-ABS${_\text{p}}$ algorithm for comparison with DCPSO-ABS. The DCPSO-ABS${_\text{p}}$ algorithm is derived from DCPSO-ABS, with the introduction of \textbf{\textit{P}} replacing \textit{\textbf{Q}}. Furthermore, to elucidate the impact of the D-C on exploration and exploitation, we employ two variants of the DCPSO-ABS algorithm, denoted as PSO-ABS${_\text{non-G-channel}}$, PSO-ABS${_\text{G-channel}}$, for comparison with PRGPSO-ABS. It is important to note that PSO-ABS${_\text{non-G-channel}}$ signifies the exclusive participation of the \textit{ non-G-channel } in the algorithm, whereas  PSO-ABS${_\text{G-channel}}$ indicates the sole involvement of the \textit{G-channel}. The experimental results, quantified by mean error and rank, are presented in Table \ref{strategy 1+2 influence performance}, where instances of the Best-Mean and Second-Mean are indicated in red and blue, respectively. Additionally, the corresponding performance curves  are plotted in Fig. \ref{Performance of PRGPSO$_{exploration}$, PRGPSO$_{exploitation}$ and PRGPSO-ABS on (a) F$_1$,   (b) F$_5$, (c) F$_{51}$, (d) F$_{57}$.}. 

In Table \ref{strategy 1+2 influence performance},  considering $F_1$-$F_7$ (Unimodal Functions),  and by comparing DCPSO-ABS,  PSO-ABS${_\text{G-channel}}$ and  PSO-ABS${_\text{non-G-channel}}$, it is obvious that  PSO-ABS${_\text{G-channel}}$ illustrates significantly better performance than PSO-ABS${_\text{non-G-channel}}$, underscoring the \textit{G-channel} holds a stronger exploitation ability  than \textit{non-G-channel}, consistent with the theoretical investigation in Section \ref{DCPSO-ABS}. Moreover, DCPSO-ABS shows the approximate performance with PSO-ABS${_\text{G-channel}}$, validating that DCPSO-ABS can maintain exploitation ability. Furthermore, DCPSO-ABS${_\text{p}}$ shows the worst performance, which is because of the low  activity of the algorithm caused by the absence of \textit{\textbf{P}}'s exploration ability. It implies the indispensability of PDG for DCPSO-ABS. Further considering  $F_{40}$-$F_{57}$ (Complex Multimodal Functions), we can obsearve that PSO-ABS${_\text{G-channel}}$ has worse performance than PSO-ABS${_\text{non-G-channel}}$, intuitively illustrating the limitations of relying only on \textit{\textbf{G}}. Moreover, DCPSO-ABS performs the remarkable performance compared with PSO-ABS${_\text{non-G-channel}}$ and PSO-ABS${_\text{G-channel}}$, exhibiting the significance of \textit{non-G-channel} and\textit{ G-channel} and the indispensability of D-C. Compared with  DCPSO-ABS${_\text{p}}$,  DCPSO-ABS performs better, further indicating the indispensability of PDG. Additionally, Fig. \ref{Performance of PRGPSO$_{exploration}$, PRGPSO$_{exploitation}$ and PRGPSO-ABS on (a) F$_1$,   (b) F$_5$, (c) F$_{51}$, (d) F$_{57}$.} (a) and (b) show that DCPSO-ABS possesses almost the same convergence speed as PSO-ABS${_\text{G-channel}}$. It further shows that the significance of the convergence speed of \textit{\textbf{G}} and that our proposed algorithm can  maintain this feature of \textit{\textbf{G}}. Besides, from  Fig. \ref{Performance of PRGPSO$_{exploration}$, PRGPSO$_{exploitation}$ and PRGPSO-ABS on (a) F$_1$,   (b) F$_5$, (c) F$_{51}$, (d) F$_{57}$.} (c) and (d), the DCPSO-ABS displays stronger convergence performance than the other compared algorithms.  These experimental results further demonstrate the  indispensability of D-C and PDG for DCPSO-ABS. In summary, the combination of the D-C and PDG indeed aids in effectively regulating the behaviors of \textbf{\textit{P}} and \textbf{\textit{G}}, thereby effectively  achieving a harmonious balance between \textit{Ei} and \textit{Er}.

%

\begin{figure}
\centering
		\includegraphics[width=3.3in]{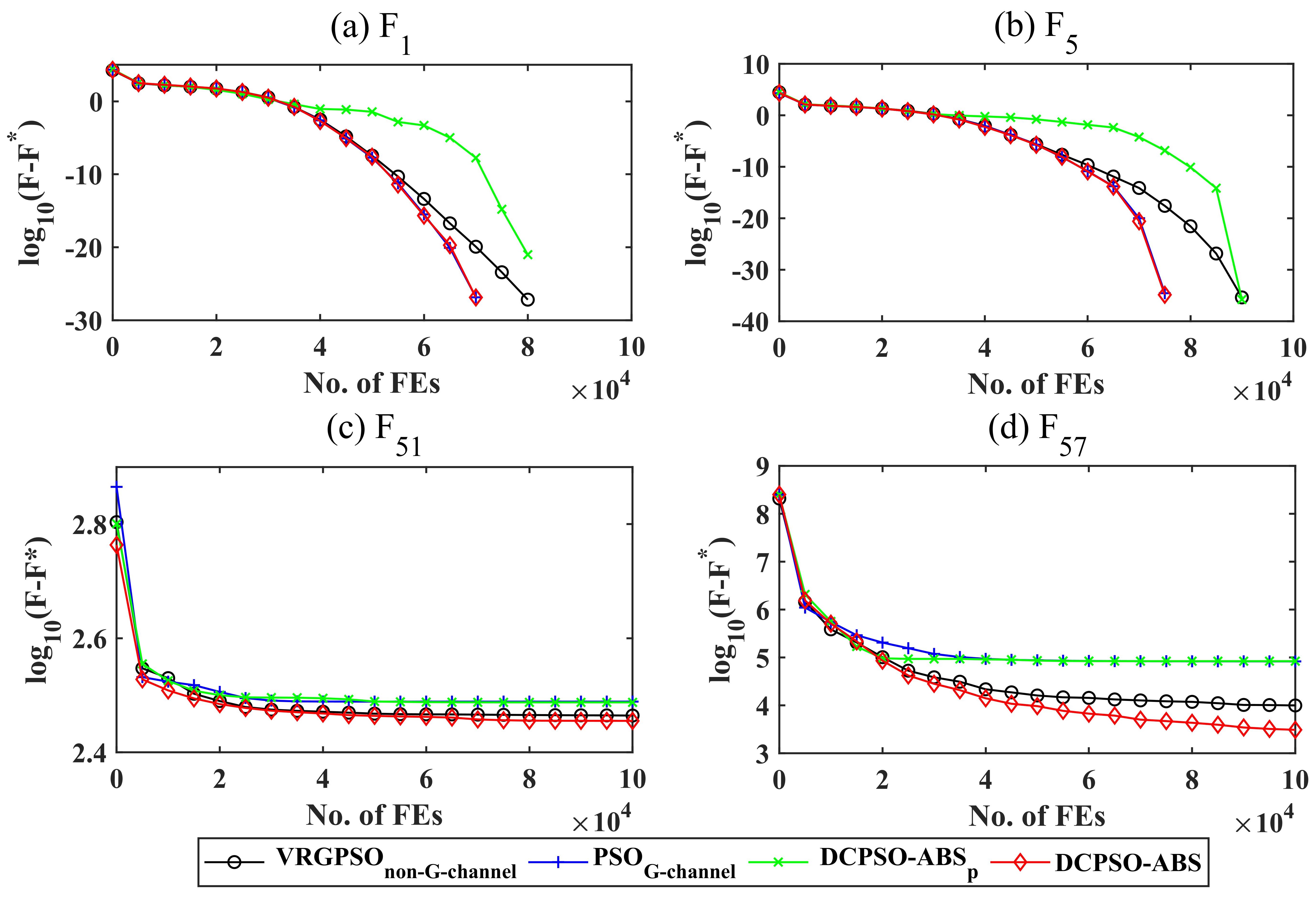}\\
	\caption{Performance curves of DCPSO-ABS${_\text{p}}$,PSO-ABS${_\text{non-G-channel}}$, PSO-ABS${_\text{G-channel}}$, and DCPSO-ABS on (a) F$_1$. (b) F$_5$. (c) F$_{51}$. (d) F$_{57}$.}
	\label{Performance of PRGPSO$_{exploration}$, PRGPSO$_{exploitation}$ and PRGPSO-ABS on (a) F$_1$,   (b) F$_5$, (c) F$_{51}$, (d) F$_{57}$.}
\end{figure}

	\section{Conclusion}\label{conclusion}
	In this paper, we investigated two behaviors of \textit{\textbf{P}} and \textit{\textbf{G}} and illustrated the mechanisms by which the behaviors of \textit{\textbf{P}} and \textit{\textbf{G}} affect the balance between \textit{Er} and \textit{Ei}.  By tracing the particle positions and potential search regions, we illustrated the reasons why the behavior of \textit{\textbf{P}} exists lie in the influence of \textit{\textbf{G}} on \textit{\textbf{P}}, the exclusive reliance of \textit{\textbf{P}} on information from its corresponding particle, and the uncontrollable number of search region searches. We also  elaborated that the existence of \textit{\textbf{G}}'s  behavior derives from the absolute reliance of each particle on \textit{\textbf{G}}.
	
	Then, we proposed a variant of PSO, DCPSO-ABS to manage the two behaviors to achieve a better balance of \textit{Er} and \textit{Ei}. In DCPSO-ABS, we constructed a DC framework including two distinct channels and two classes of particles, for mitigating the influence of \textit{\textbf{G}} on \textit{\textbf{P}} and breaking the status quo of each particle’s absolute reliance on \textit{\textbf{G}} in standard PSO. We considered the other \textit{\textbf{$\textbf{P}_n$}}s for introducing new direction elements to the current \textit{\textbf{P}}, connected the updated statuses of particles with the quality of the current search region for controlling the number of searches within the current region,  and  established the mathematic relationship between the number of function evaluations and the proportion of the two iterative channels employed, consequently, proposed ABS strategy for management two behaviors of \textit{\textbf{P}} and \textit{\textbf{G}}. 
	
	The experimental results verified the indispensability of DC framework and ABS strategy for balancing between \textit{Er} and \textit{\textit{Ei}}, the convergence performance of our proposed algorithm on different problems, and its stability and scalability for different numbers of function evaluations and dimensions, justifiably demonstrating the strong generalization performance of DCPSO-ABS. Therefore, in the future, we would like to apply our idea to other population-based optimization algorithms for the balance of \textit{Er} and \textit{Ei}.

%
\newpage
	\bibliographystyle{IEEEtran}
	\bibliography{Main_Manuscript}

\begin{thebibliography}{10}
\providecommand{\url}[1]{#1}
\csname url@samestyle\endcsname
\providecommand{\newblock}{\relax}
\providecommand{\bibinfo}[2]{#2}
\providecommand{\BIBentrySTDinterwordspacing}{\spaceskip=0pt\relax}
\providecommand{\BIBentryALTinterwordstretchfactor}{4}
\providecommand{\BIBentryALTinterwordspacing}{\spaceskip=\fontdimen2\font plus
\BIBentryALTinterwordstretchfactor\fontdimen3\font minus
  \fontdimen4\font\relax}
\providecommand{\BIBforeignlanguage}[2]{{%
\expandafter\ifx\csname l@#1\endcsname\relax
\typeout{** WARNING: IEEEtran.bst: No hyphenation pattern has been}%
\typeout{** loaded for the language `#1'. Using the pattern for}%
\typeout{** the default language instead.}%
\else
\language=\csname l@#1\endcsname
\fi
#2}}
\providecommand{\BIBdecl}{\relax}
\BIBdecl

\bibitem{kennedy1995particle}
J.~Kennedy and R.~Eberhart, ``Particle swarm optimization,'' in \emph{Proc.
  ICNN'95-international Conf. neural Netw.}, vol.~4, 1995, pp. 1942--1948.

\bibitem{katoch2021review}
S.~Katoch, S.~S. Chauhan, and V.~Kumar, ``A review on genetic algorithm: past,
  present, and future,'' \emph{Multimedia Tools Appl.}, vol.~80, no.~5, pp.
  8091--8126, 2021.

\bibitem{zhou2019self}
S.~Zhou, L.~Xing, X.~Zheng, N.~Du, L.~Wang, and Q.~Zhang, ``A self-adaptive
  differential evolution algorithm for scheduling a single batch-processing
  machine with arbitrary job sizes and release times,'' \emph{IEEE Trans.
  Cybern.}, vol.~51, no.~3, pp. 1430--1442, 2019.

\bibitem{lynn2015heterogeneous}
N.~Lynn and P.~N. Suganthan, ``Heterogeneous comprehensive learning particle
  swarm optimization with enhanced exploration and exploitation,'' \emph{Swarm
  Evol. Comput.}, vol.~24, pp. 11--24, 2015.

\bibitem{houssein2021major}
E.~H. Houssein, A.~G. Gad, K.~Hussain, and P.~N. Suganthan, ``Major advances in
  particle swarm optimization: theory, analysis, and application,'' \emph{Swarm
  Evol. Comput.}, vol.~63, p. 100868, 2021.

\bibitem{isho2020persistence}
B.~Isho, K.~T. Abe, M.~Zuo, A.~J. Jamal, B.~Rathod, J.~H. Wang, Z.~Li, G.~Chao,
  O.~L. Rojas, Y.~M. Bang \emph{et~al.}, ``Persistence of serum and saliva
  antibody responses to sars-cov-2 spike antigens in covid-19 patients,''
  \emph{Sci. immunology}, vol.~5, no.~52, p. eabe5511, 2020.

\bibitem{shi1999empirical}
Y.~Shi and R.~C. Eberhart, ``Empirical study of particle swarm optimization,''
  in \emph{Proc. 1999 Congr. Evol. computation-CEC99 Cat. No. 99TH8406},
  vol.~3, 1999, pp. 1945--1950.

\bibitem{ratnaweera2004self}
A.~Ratnaweera, S.~K. Halgamuge, and H.~C. Watson, ``Self-organizing
  hierarchical particle swarm optimizer with time-varying acceleration
  coefficients,'' \emph{IEEE Trans. Evol. Comput.}, vol.~8, no.~3, pp.
  240--255, 2004.

\bibitem{liu2019novel}
W.~Liu, Z.~Wang, Y.~Yuan, N.~Zeng, K.~Hone, and X.~Liu, ``A novel
  sigmoid-function-based adaptive weighted particle swarm optimizer,''
  \emph{IEEE Trans. Cybern.}, vol.~51, no.~2, pp. 1085--1093, 2019.

\bibitem{qu2012distance}
B.-Y. Qu, P.~N. Suganthan, and S.~Das, ``A distance-based locally informed
  particle swarm model for multimodal optimization,'' \emph{IEEE Trans. Evol.
  Comput.}, vol.~17, no.~3, pp. 387--402, 2012.

\bibitem{zeng2020dynamic}
N.~Zeng, Z.~Wang, W.~Liu, H.~Zhang, K.~Hone, and X.~Liu, ``A dynamic
  neighborhood-based switching particle swarm optimization algorithm,''
  \emph{IEEE Trans. Cybern.}, vol.~52, no.~9, pp. 9290--9301, 2020.

\bibitem{huang2012example}
H.~Huang, H.~Qin, Z.~Hao, and A.~Lim, ``Example-based learning particle swarm
  optimization for continuous optimization,'' \emph{Inf. Sci.}, vol. 182,
  no.~1, pp. 125--138, 2012.

\bibitem{wang2018hybrid}
F.~Wang, H.~Zhang, K.~Li, Z.~Lin, J.~Yang, and X.-L. Shen, ``A hybrid particle
  swarm optimization algorithm using adaptive learning strategy,'' \emph{Inf.
  Sci.}, vol. 436, pp. 162--177, 2018.

\bibitem{ren2013scatter}
Z.~Ren, A.~Zhang, C.~Wen, and Z.~Feng, ``A scatter learning particle swarm
  optimization algorithm for multimodal problems,'' \emph{IEEE Trans. Cybern.},
  vol.~44, no.~7, pp. 1127--1140, 2013.

\bibitem{shaheen2021novel}
M.~A. Shaheen, H.~M. Hasanien, and A.~Alkuhayli, ``A novel hybrid gwo-pso
  optimization technique for optimal reactive power dispatch problem
  solution,'' \emph{Ain Shams Eng. J.}, vol.~12, no.~1, pp. 621--630, 2021.

\bibitem{7271066}
Y.-J. Gong, J.-J. Li, Y.~Zhou, Y.~Li, H.~S.-H. Chung, Y.-H. Shi, and J.~Zhang,
  ``Genetic learning particle swarm optimization,'' \emph{IEEE Trans. Cybern.},
  vol.~46, no.~10, pp. 2277--2290, 2016.

\bibitem{shi1998modified}
Y.~Shi and R.~Eberhart, ``A modified particle swarm optimizer,'' in \emph{1998
  IEEE Int. Conf. Evol. Comput. proceedings. IEEE world Congr. Comput. Intell.
  Cat. No. 98TH8360}, 1998, pp. 69--73.

\bibitem{nickabadi2011novel}
A.~Nickabadi, M.~M. Ebadzadeh, and R.~Safabakhsh, ``A novel particle swarm
  optimization algorithm with adaptive inertia weight,'' \emph{Appl. soft
  Comput.}, vol.~11, no.~4, pp. 3658--3670, 2011.

\bibitem{taherkhani2016novel}
M.~Taherkhani and R.~Safabakhsh, ``A novel stability-based adaptive inertia
  weight for particle swarm optimization,'' \emph{Appl. Soft Comput.}, vol.~38,
  pp. 281--295, 2016.

\bibitem{wei2020multiple}
B.~Wei, X.~Xia, F.~Yu, Y.~Zhang, X.~Xu, H.~Wu, L.~Gui, and G.~He, ``Multiple
  adaptive strategies based particle swarm optimization algorithm,''
  \emph{Swarm Evol. Comput.}, vol.~57, p. 100731, 2020.

\bibitem{moazen2023pso}
H.~Moazen, S.~Molaei, L.~Farzinvash, and M.~Sabaei, ``Pso-elpm: Pso with elite
  learning, enhanced parameter updating, and exponential mutation operator,''
  \emph{Inf. Sci.}, vol. 628, pp. 70--91, 2023.

\bibitem{chen2012particle}
W.-N. Chen, J.~Zhang, Y.~Lin, N.~Chen, Z.-H. Zhan, H.~S.-H. Chung, Y.~Li, and
  Y.-H. Shi, ``Particle swarm optimization with an aging leader and
  challengers,'' \emph{IEEE Trans. Evol. Comput.}, vol.~17, no.~2, pp.
  241--258, 2012.

\bibitem{zhan2009adaptive}
Z.-H. Zhan, J.~Zhang, Y.~Li, and H.~S.-H. Chung, ``Adaptive particle swarm
  optimization,'' \emph{IEEE Trans. Syst., Man, Cybern., B Cybern.}, vol.~39,
  no.~6, pp. 1362--1381, 2009.

\bibitem{kennedy2002population}
J.~Kennedy and R.~Mendes, ``Population structure and particle swarm
  performance,'' in \emph{Proc. 2002 Congr. Evol. Computation. CEC'02 Cat. No.
  02TH8600}, vol.~2, 2002, pp. 1671--1676.

\bibitem{mendes2004fully}
R.~Mendes, J.~Kennedy, and J.~Neves, ``The fully informed particle swarm:
  simpler, maybe better,'' \emph{IEEE Trans. Evol. Comput.}, vol.~8, no.~3, pp.
  204--210, 2004.

\bibitem{zhang2011scale}
C.~Zhang and Z.~Yi, ``Scale-free fully informed particle swarm optimization
  algorithm,'' \emph{Inf. Sci.}, vol. 181, no.~20, pp. 4550--4568, 2011.

\bibitem{zhang2021promotive}
L.~Zhang, S.-K. Oh, W.~Pedrycz, B.~Yang, and L.~Wang, ``A promotive particle
  swarm optimizer with double hierarchical structures,'' \emph{IEEE Trans.
  Cybern.}, vol.~52, no.~12, pp. 13\,308--13\,322, 2021.

\bibitem{zhan2009orthogonal}
Z.-H. Zhan, J.~Zhang, and O.~Liu, ``Orthogonal learning particle swarm
  optimization,'' in \emph{Proc. 11th Annu. Conf. Genetic Evol. Comput.}, 2009,
  pp. 1763--1764.

\bibitem{liang2006comprehensive}
J.~J. Liang, A.~K. Qin, P.~N. Suganthan, and S.~Baskar, ``Comprehensive
  learning particle swarm optimizer for global optimization of multimodal
  functions,'' \emph{IEEE Trans. Evol. Comput.}, vol.~10, no.~3, pp. 281--295,
  2006.

\bibitem{li2015composite}
J.~Li, J.~Zhang, C.~Jiang, and M.~Zhou, ``Composite particle swarm optimizer
  with historical memory for function optimization,'' \emph{IEEE Trans.
  Cybern.}, vol.~45, no.~10, pp. 2350--2363, 2015.

\bibitem{van2004cooperative}
F.~Van~den Bergh and A.~P. Engelbrecht, ``A cooperative approach to particle
  swarm optimization,'' \emph{IEEE Trans. Evol. Comput.}, vol.~8, no.~3, pp.
  225--239, 2004.

\bibitem{li2015competitive}
Y.~Li, Z.-H. Zhan, S.~Lin, J.~Zhang, and X.~Luo, ``Competitive and cooperative
  particle swarm optimization with information sharing mechanism for global
  optimization problems,'' \emph{Inf. Sci.}, vol. 293, pp. 370--382, 2015.

\bibitem{niu2007mcpso}
B.~Niu, Y.~Zhu, X.~He, and H.~Wu, ``Mcpso: A multi-swarm cooperative particle
  swarm optimizer,'' \emph{Appl. Math. Comput.}, vol. 185, no.~2, pp.
  1050--1062, 2007.

\bibitem{li2011hybrid}
S.~Li, M.~Tan, I.~W. Tsang, and J.~T.-Y. Kwok, ``A hybrid pso-bfgs strategy for
  global optimization of multimodal functions,'' \emph{IEEE Trans. Syst., Man,
  Cybern., B Cybern.}, vol.~41, no.~4, pp. 1003--1014, 2011.

\bibitem{cao2018comprehensive}
Y.~Cao, H.~Zhang, W.~Li, M.~Zhou, Y.~Zhang, and W.~A. Chaovalitwongse,
  ``Comprehensive learning particle swarm optimization algorithm with local
  search for multimodal functions,'' \emph{IEEE Trans. Evol. Comput.}, vol.~23,
  no.~4, pp. 718--731, 2018.

\bibitem{fan2004hybrid}
S.-K.~S. Fan, Y.-c. Liang, and E.~Zahara, ``Hybrid simplex search and particle
  swarm optimization for the global optimization of multimodal functions,''
  \emph{Eng. Optim.}, vol.~36, no.~4, pp. 401--418, 2004.

\bibitem{hakli2014novel}
H.~Hakl{\i} and H.~U{\u{g}}uz, ``A novel particle swarm optimization algorithm
  with levy flight,'' \emph{Appl. Soft Comput.}, vol.~23, pp. 333--345, 2014.

\bibitem{farnad2018new}
B.~Farnad, A.~Jafarian, and D.~Baleanu, ``A new hybrid algorithm for continuous
  optimization problem,'' \emph{Appl. Math. Modelling}, vol.~55, pp. 652--673,
  2018.

\bibitem{lynn2017ensemble}
N.~Lynn and P.~N. Suganthan, ``Ensemble particle swarm optimizer,'' \emph{Appl.
  Soft Comput.}, vol.~55, pp. 533--548, 2017.

\bibitem{peram2003fitness}
T.~Peram, K.~Veeramachaneni, and C.~K. Mohan, ``Fitness-distance-ratio based
  particle swarm optimization,'' in \emph{Proc. 2003 IEEE Swarm Intell.
  Symposium. SIS'03 Cat. No. 03EX706}, 2003, pp. 174--181.

\bibitem{scholey2003cell}
J.~M. Scholey, I.~Brust-Mascher, and A.~Mogilner, ``Cell division,''
  \emph{Nature}, vol. 422, no. 6933, pp. 746--752, 2003.

\bibitem{olorunda2008measuring}
O.~Olorunda and A.~P. Engelbrecht, ``Measuring exploration/exploitation in
  particle swarms using swarm diversity,'' in \emph{2008 IEEE Congr. Evol.
  Comput. IEEE world Congr. Comput. Intell.}, 2008, pp. 1128--1134.

\bibitem{liang2013problem}
J.~J. Liang, B.~Qu, P.~N. Suganthan, and A.~G. Hern{\'a}ndez-D{\'\i}az,
  ``Problem definitions and evaluation criteria for the cec 2013 special
  session on real-parameter optimization,'' \emph{Comput. Intell. Lab.,
  Zhengzhou Univ., Zhengzhou, China Nanyang Technological Univ., Singap., Tech.
  Rep.}, vol. 201212, no.~34, pp. 281--295, 2013.

\bibitem{wu2017problem}
G.~Wu, R.~Mallipeddi, and P.~N. Suganthan, ``Problem definitions and evaluation
  criteria for the cec 2017 competition on constrained real-parameter
  optimization,'' \emph{Nat. Univ. Defense Technol., Changsha, Hunan, PR China
  Kyungpook Nat. Univ., Daegu, South Korea Nanyang Technological Univ.,
  Singap., Tech. Rep.}, 2017.

\bibitem{wolpert1997no}
D.~H. Wolpert and W.~G. Macready, ``No free lunch theorems for optimization,''
  \emph{IEEE Trans. Evol. Comput.}, vol.~1, no.~1, pp. 67--82, 1997.

\end{thebibliography}

	\vfill
	
\end{document}